\documentclass{article}

\PassOptionsToPackage{numbers, compress}{natbib}
\usepackage[final, dandb]{neurips_2025}

\usepackage[utf8]{inputenc} % allow utf-8 input
\usepackage[T1]{fontenc}    % use 8-bit T1 fonts
\usepackage{hyperref}       % hyperlinks
\usepackage{url}            % simple URL typesetting
\usepackage{booktabs}       % professional-quality tables
\usepackage{amsfonts}       % blackboard math symbols
\usepackage{nicefrac}       % compact symbols for 1/2, etc.
\usepackage{microtype}      % microtypography
\usepackage{xcolor}         % colors
\usepackage{amsmath}
\usepackage{stmaryrd}
\usepackage{cleveref}

\usepackage[utf8]{inputenc} % allow utf-8 input
\usepackage[LGR, T1]{fontenc}
\usepackage{hyperref}       % hyperlinks
\usepackage{url}            % simple URL typesetting
\usepackage{booktabs}       % professional-quality tables
\usepackage{amsfonts}       % blackboard math symbols
\usepackage{nicefrac}       % compact symbols for 1/2, etc.
\usepackage{microtype}      % microtypography
\usepackage{xcolor}         % colors
\usepackage{makecell}
\usepackage[framemethod=TikZ]{mdframed}
\usepackage{amssymb}% http://ctan.org/pkg/amssymb
\usepackage{pifont}% http://ctan.org/pkg/pifont

\usepackage{enumitem}
\usepackage{xspace}

\usepackage{minted}
\usepackage{wrapfig}
\usepackage{clrscode}
\usepackage{subcaption}
\usepackage{inconsolata}

\usepackage{color}
\definecolor{keywordcolor}{rgb}{0.7, 0.1, 0.1}   % red
\definecolor{tacticcolor}{rgb}{0.0, 0.1, 0.6}    % blue
\definecolor{commentcolor}{rgb}{0.4, 0.4, 0.4}   % grey
\definecolor{symbolcolor}{rgb}{0.0, 0.1, 0.6}    % blue
\definecolor{sortcolor}{rgb}{0.1, 0.5, 0.1}      % green
\definecolor{attributecolor}{rgb}{0.7, 0.1, 0.1} % red

\usepackage{listings}

\renewcommand{\paragraph}[1]{\noindent \textbf{#1}}

\DeclareUnicodeCharacter{2223}{$\mid$}
\DeclareUnicodeCharacter{211D}{$\mathbb{R}$}
\DeclareUnicodeCharacter{2081}{$_{1}$}
\DeclareUnicodeCharacter{2082}{$_{2}$}
\DeclareUnicodeCharacter{22A2}{$\vdash$}
\DeclareUnicodeCharacter{2115}{$\mathbb{N}$}
\DeclareUnicodeCharacter{2211}{$\sum$}
\DeclareUnicodeCharacter{2194}{$\leftrightarrow$}
\DeclareUnicodeCharacter{2208}{$\in$}
\DeclareUnicodeCharacter{1D55C}{$\mathbb{k}$}
\DeclareUnicodeCharacter{3B9}{$\iota$}
\DeclareUnicodeCharacter{3C0}{$\pi$}
\DeclareUnicodeCharacter{271D}{$\dagger$}
\DeclareUnicodeCharacter{2075}{$^{5}$}
\DeclareUnicodeCharacter{2074}{$^{4}$}
\DeclareUnicodeCharacter{2073}{$^{3}$}
\DeclareUnicodeCharacter{2072}{$^{2}$}
\DeclareUnicodeCharacter{2071}{$^{1}$}
\DeclareUnicodeCharacter{2200}{$\forall$}
\DeclareUnicodeCharacter{2A0D}{$\int$}
\usepackage[textsize=scriptsize]{todonotes}
\usepackage{cleveref}

\newcommand{\clever}{\textsc{Clever}\xspace}

\newcommand{\name}{\clever}
\newcommand{\humaneval}{\textsc{HumanEval}\xspace}

\title{\name: A Curated Benchmark for Formally Verified Code Generation}

\author{%
  Amitayush Thakur$^{\dagger}$ \\
  \texttt{amitayush@utexas.edu}
  \And
  Jasper Lee$^{\dagger}$ \\
  \texttt{leejasper@utexas.edu}
  \And
  George Tsoukalas$^{\dagger}$ \\
  \texttt{george.tsoukalas@utexas.edu}
  \And
  Meghana Sistla$^{\dagger}$ \\
  \texttt{mesistla@utexas.edu}
  \And
  Matthew Zhao$^{\dagger}$ \\
  \texttt{matthewzhao@utexas.edu}
  \AND % New line for the next row of authors
  Stefan Zetzsche$^{\ddagger}$ \\
  \texttt{stefanze@amazon.co.uk}
  \And
  Greg Durrett$^{\dagger}$ \\
  \texttt{gdurrett@cs.utexas.edu}
  \And
  Yisong Yue$^{\star}$ \\
  \texttt{yyue@caltech.edu}
  \And
  Swarat Chaudhuri$^{\dagger}$ \\
  \texttt{swarat@cs.utexas.edu}\\ \\
  $^{\dagger}$ UT Austin \quad
  $^{\ddagger}$ Amazon \quad
  $^{\star}$ Caltech
}

\begin{document}

\maketitle

\begin{abstract}
We introduce \clever\footnote{CLEVER: Curated Lean Verified Code Generation Benchmark}, a high-quality, curated benchmark of 161 problems for end-to-end verified code generation in Lean. 
Each problem consists of (1) the task of generating a specification that matches a held-out ground-truth specification, and (2) the task of generating a Lean implementation that provably satisfies this specification. Unlike prior benchmarks, \clever avoids test-case supervision, LLM-generated annotations, and specifications that leak implementation logic or allow vacuous solutions. All outputs are verified post-hoc using Lean's type checker to ensure machine-checkable correctness.  We use \clever to evaluate several few-shot and agentic approaches based on state-of-the-art language models.
These methods all struggle to achieve full verification, 
establishing it as a challenging frontier benchmark for program synthesis and formal reasoning. Our benchmark can be found on \href{https://github.com/trishullab/clever}{\textcolor{blue}{GitHub}} as well as \href{https://huggingface.co/datasets/amitayusht/clever}{\textcolor{blue}{HuggingFace}}. All our evaluation code is also available \href{https://github.com/trishullab/clever-prover}{\textcolor{blue}{online}}.
\end{abstract}

\section{Introduction}
% --- Define macros for symbols ---
\newcommand{\SpecLogicRv}{\mathsf{\Psi}}
\newcommand{\TestSuiteRv}{\mathsf{T}}
\newcommand{\ProgramRv}{\mathsf{P}}
\newcommand{\ProgramSet}{\Pi}
\newcommand{\SpecNLRv}{\mathsf{S_{NL}}}
\newcommand{\SpecNLSet}{\mathcal{N}}
\newcommand{\SpecF}{\psi}
\newcommand{\SpecFFull}{\phi(x_1, \dots, x_n; \Program)}
\newcommand{\SpecNL}{\nu}
\newcommand{\SpecFSet}{\Psi}
\newcommand{\Program}{\pi}
\newcommand{\TestCase}{\tau}
\newcommand{\TestSet}{\mathcal{T}}

Interactive theorem-provers (ITPs) \citep{huet1997coq,paulson1994isabelle,de2015lean} are an established technology for engineering high-assurance software, leading to success stories like the CompCert verified C compiler \citep{leroy2009formal} and the seL4 \citep{klein2009sel4} verified microkernel.
However, writing formal specifications and correctness proofs for software systems can take tremendous effort --- for example, the development of seL4 was reported to take 20+ person-years. These costs are a key impediment to the broad deployment of ITP-based formal verification. 

Recent progress in autoformalization and neural theorem-proving~\citep{polu2020generative, li2024surveydeeplearningtheorem} has raised hopes of scaling up formal verification \citep{yang2024formal}. 
Most existing work in this area has focused on formalizing and proving statements in pure mathematics \cite{zheng2021minif2f,tsoukalas2024putnambenchevaluatingneuraltheoremprovers}. 
However, the software verification setting opens up the challenge of \emph{generating code that is formally verified by construction}, a problem without a well-studied analog in the mathematics setting. 

To date, there are a handful of benchmarks \citep{dougherty2025provingcodinginterviewbenchmark,loughridge2024dafnybenchbenchmarkformalsoftware, lohn2024minicodepropsminimalbenchmarkproving} for formally verified code generation.
%\stefan{Specify that you mean verified code generation in Lean. There are more verified code generation benchmarks for other languages (which you might want to cite too?).} 
However, 
%as we show in Appendix~\Cref{app:fvapps}, the formal specifications in F do not always 
the formal specifications in these benchmarks tend not to capture the full (natural-language) intent behind the target program and sometimes hint at ways to implement the program. This ambiguity allows a code generator to ``cheat'' by generating trivial programs or copying code from the specification (see \Cref{sec:fvapps}). 
%uses test cases to check the correctness of the generated code, which only provides partial validation.

%allowing the code generator to ``cheat'' by synthesizing trivial programs.  
%Second \emph{leaky specifications} ---  (see \Cref{fig:leak-issue-examples}); and 
%(3) \emph{reliance on test cases}, which provide only partial validation.

In this paper, we address this gap in the prior art with \clever, a high-quality benchmark for formally verified AI-based code generation. \clever includes hand-crafted Lean specifications of 161 programming tasks from the \humaneval benchmark \citep{chen2021evaluating}.

%The goal here is to synthesize, based on a natural-language problem specification, a formal specification and a program that implements the specification, along with Lean proofs of both specification soundness and implementation correctness. 
%Unlike prior work, our benchmark enforces complete specification coverage and proof-backed correctness, avoiding shortcut strategies enabled by test-case-based evaluation.
%We present a high-quality benchmark suite for \humaneval-style~\cite{chen2021evaluating} problems in Lean~4 that enables rigorous evaluation of end-to-end verified code generation from natural language. 
It evaluates models in two stages: (1) \textit{Specification certification:} Given a natural language specification, the model is required to generate a Lean specification and prove that it is semantically equivalent to the ground-truth specification. (2) \textit{Implementation certification:} Once the model has correctly generated the specification, it is required to generate a Lean implementation and prove that it satisfies the ground-truth specification. A synthesis attempt is deemed \emph{successful} only when both the proofs generated in the two stages are fully verified by Lean's type checker. This rigorous pipeline avoids the pitfalls of both automatically generated specifications and test-based supervision.

We use \clever to evaluate several state-of-the-art LLMs prompted in a few-shot manner and show that they can only solve up to  \texttt{1/161} end-to-end verified code generation problem, establishing \clever as a challenging frontier benchmark for program synthesis and formal reasoning.
%We specifically avoid informal test-based supervision and requiring full machine-checkable correctness guarantees. 
In summary, our contributions include:
\begin{enumerate}[leftmargin=0.2in]
\item We introduce \clever, the first curated benchmark for evaluating the generation of specifications and formally verified code in Lean. 
%\stefan{I think this claim is too strong. E.g., https://github.com/JetBrains-Research/HumanEval-Dafny is hand-curated too.}
% \gd{Say ``The benchmark evaluates'' instead? Also put the number of instances in the benchmark somewhere here}
The benchmark comprises of 161 programming problems; it evaluates both \textit{formal specification generation} and \textit{implementation synthesis} from natural language, requiring formal correctness proofs for both. 
%Each benchmark instance is solved via two tasks: (1) \emph{spec certification}, where the generated formal specification \(\SpecF\) must be proven equivalent to the ground-truth~\(\SpecF^*\); and (2) \emph{implementation certification}, where the generated program~\(\Program\) must provably satisfy~\(\SpecF^*\). Success requires both Lean 4 code and its associated proofs to type-check.
%\item \textbf{Leak-proof, non-computable specifications.}
All specifications are manually written to be complete, implementation-agnostic, and free from exploitable artifacts, preventing models from shortcutting the intended semantics.

\item We present an empirical evaluation of several state-of-the-art LLMs and agentic approaches on \clever and show that they all struggle at meeting the benchmark's goals, establishing the challenging nature of the benchmark.  
%\item 
%We introduce a new proof synthesis approach that leverages proof planning and auxiliary lemma synthesis to solve complex verification tasks that are beyond the capabilities of baseline methods.
\end{enumerate}

\begin{figure*}
    \centering
    \includegraphics[scale=0.4]{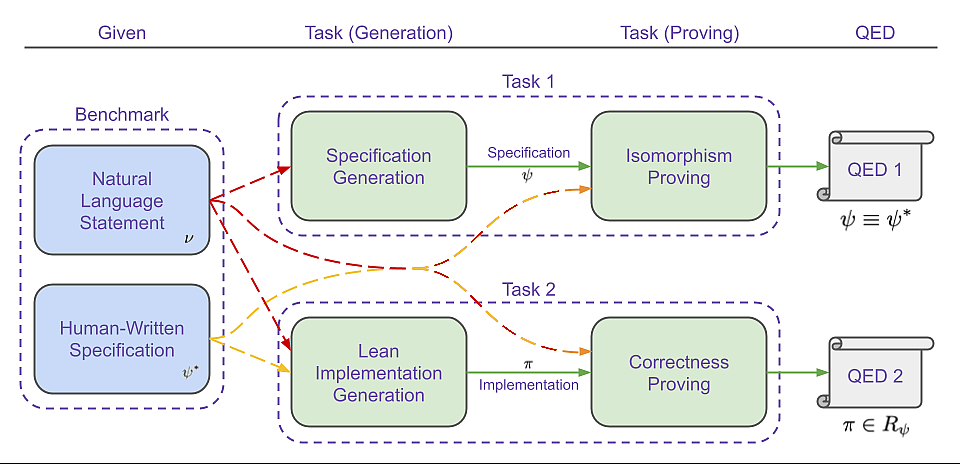}
    \caption{The two tasks of the \clever benchmark pipeline. Task 1 requires first generating a specification $\psi$ from the natural language statement $\nu$, then proving an isomorphism between the generated specification and a human-written specification $\psi^*$. Task 2 requires first generating a Lean implementation $\pi$, then proving its correctness according to the human-written specification. Both of these tasks must be completed correctly (reaching both QED 1 and QED 2) in order for a success to be counted.}
    \label{fig:clever-summary}
    \vspace{-0.2in}
\end{figure*}
% \george{Nit: can we pass figure 1 through an image sharpener? it is a little blurry}

\section{The \clever Benchmark}
% \label{sec:problem-formulation}
\label{sec:problem-formulation}

\clever builds on \humaneval~\citep{chen2021evaluating} by adapting 161\footnote{Not all problems could be formalized due to limitations in Lean 4 and its supported libraries.} of its 164 programming problems for formal verification in Lean 4. Each problem includes a natural language description (\(\SpecNL\)), a human-authored formal specification (\(\SpecF^*\)), a Lean function signature (\(\Program_{\text{sig}}\)) for the implementation, and Lean theorems for both specification equivalence and implementation correctness. All formal specifications are written as \emph{non-computable} logical propositions --- i.e., they use quantifiers and logical connectives that cannot be directly evaluated --- ensuring that models cannot copy implementation logic from specification syntax.

During evaluation, a model being evaluated on the benchmark starts with the natural-language description \(\SpecNL\). Given this text, the model must generate:
\begin{enumerate}[label=(\arabic*),leftmargin=0.3in]
    \item a formal Lean specification \(\SpecF\), expressed as a predicate (a function that returns a Lean 4 proposition i.e. \texttt{Prop}),
    \item a proof that \(\SpecF\) is semantically equivalent to a hidden ground-truth Lean specification \(\SpecF^*\),
    \item a Lean implementation\footnote{Here, we use the fact that Lean is not just a language for mathematical specifications and proofs but a full-fledged functional programming language.} \(\Program\) that matches the function signature (\(\Program_{\text{sig}}\)) and is designed to satisfy \(\SpecF^*\) (and hence \(\SpecF\)), and
    \item a formal proof establishing that \(\Program\) satisfies \(\SpecF^*\).
\end{enumerate}

These steps (\Cref{fig:clever-summary}) form two certification goals: (1) \textit{Specification certification:} Steps 1–2 verify that the model correctly inferred the intended behavior. (2) \textit{Implementation certification:} Steps 3–4 verify that the generated implementation satisfies the formal intent. 

%Importantly, the ground-truth specifications in \clever are carefully crafted to be not computable, thus avoiding leakage from specifications to implementations. 
Our staged reasoning setup allows fine-grained diagnosis: models may fail at generating specifications, proving equivalence between the generated and ground-truth specifications, synthesizing implementations, or proving implementation correctness. 
For example, note that we require the generated implementation $\pi$ to satisfy the ground-truth specification \(\SpecF^*\) instead of the model-generated specification \(\SpecF\). This is because we want the evaluation of $\pi$ to be independent of the ability of the model to generate the correct specification. More generally, failures at the various stages of our pipeline are independently diagnosed using Lean’s type checker.

\label{sec:benchmark-details}

% As mentioned in \Cref{sec:problem-formulation}\stefan{This is referring to the current section}, the benchmark defines two verification tasks: (1) \textbf{Specification Certification}—generate \(\SpecF\) from \(\SpecNL\) and prove its equivalence to a ground-truth \(\SpecF^*\); and (2) \textbf{Implementation Certification}—synthesize \(\Program\) and prove that it satisfies \(\SpecF^*\).

\begin{figure}
    \centering
    %\vspace{-0.1in}
    \begin{mdframed}[roundcorner=10pt]
    \begin{minipage}[t]{0.5\linewidth}
    (a)
    \begin{lstlisting}[basicstyle=\scriptsize\ttfamily]
-- computable spec
def problem_spec
-- function signature
(implementation: Nat → Nat)
-- inputs
(n: Nat) : Prop :=
-- spec
let spec (result: Nat) :=
match n with
| 0 => result = 0
| 1 => result = 1
| n' + 2 => result = 
implementation n' + 
implementation (n' + 1)
-- return value satisfies spec
∃ result, implementation n = result ∧ spec result
    \end{lstlisting}
    \end{minipage}
    % \vspace{-7.3in}
    % (b)\vspace{7.3in}
    \begin{minipage}[t]{0.5\linewidth}
    (b)
    \begin{lstlisting}[basicstyle=\scriptsize\ttfamily]
-- non-computable spec
inductive fibonacci_non_computable : ℕ → ℕ → Prop
| base0 : fibonacci_non_computable 0 0
| base1 : fibonacci_non_computable 1 1
| step  : ∀ n f₁ f₂, 
fibonacci_non_computable n f₁ →
fibonacci_non_computable (n + 1) f₂ →
fibonacci_non_computable (n + 2) (f₁ + f₂)

def problem_spec
-- function signature
(implementation: Nat → Nat)
-- inputs
(n: Nat) :=
-- spec
let spec (result: Nat) :=
  fibonacci_non_computable n result
-- program termination
∃ result,
  implementation xs = result ∧ spec result
    \end{lstlisting}
    \end{minipage}
    \end{mdframed}
    %\reducevspacebetweenfigureandcaption
    \vspace{-0.1in}
    \caption{
    Two different specs for finding the $n^{th}$ Fibonacci number. (a) shows a computable specification that \textit{leaks} the implementation; (b) shows a non-computable specification leading to no-leakage of the implementation and enforcing the model to learn the deeper logical inference.}
    \label{fig:non-computable-vs-computable}
\end{figure}

\paragraph{Challenges Encountered during Formalization.}
A key design decision in our benchmark is the use of \emph{non-computable} specifications, which are predicates or functions in Lean that return propositions (\texttt{Prop} in Lean) that cannot be evaluated or simplified (decided by Lean) through computation alone. These contrast with \emph{computable} specifications, written as executable functions or decidable predicates that Lean can reduce directly. While easier to verify, computable specs often \emph{leak} the desired logic: models can copy them into implementations and produce trivial proofs via rewriting. \Cref{fig:non-computable-vs-computable} shows the difference between a computable and a non-computable specification.

\Cref{fig:leak-issue-examples} demonstrates the importance of this contrast. The left side (a–c) shows a computable spec whose logic is mirrored exactly in the GPT-4o-generated implementation, enabling a trivial proof. On the right (d–f), the spec is non-computable and requires symbolic reasoning to prove correctness. Notably, the GPT-4o-generated implementation in (e) does not mirror the spec, and the proof fails without further reasoning. This design ensures that models must engage in deeper logical inference, not just syntactic pattern matching. By using non-computable specs across our benchmark, we eliminate leakage and enforce truly verified reasoning from models. 

\begin{figure}
\centering
%\vspace{-0.1in}
\begin{mdframed}[roundcorner=6pt, innerleftmargin=4pt, innerrightmargin=4pt, innertopmargin=2pt, innerbottommargin=2pt]
\begin{minipage}[t]{0.5\linewidth}
(a)
\begin{lstlisting}[basicstyle=\scriptsize\ttfamily]
def problem_spec
(implementation: List Int → Int → Bool)
(q: List Int) (w: Int) :=
let spec (result : Bool) :=
  result ↔ (List.Palindrome q) ∧ (List.sum q ≤ w)
∃ result, implementation q w = result ∧ spec result
\end{lstlisting}
\vspace{24pt}
(b)
\begin{lstlisting}[basicstyle=\scriptsize\ttfamily]
def implementation (q: List Int) (w: Int) : Bool :=
-- implementation generated by GPT-4o
List.Palindrome q ∧ List.sum q ≤ w
\end{lstlisting}
\vspace{16pt}
(c)
\begin{lstlisting}[basicstyle=\scriptsize\ttfamily]
theorem correctness (q: List Int) (w: Int)
: problem_spec implementation q w := by
-- proof generated by GPT-4o
unfold problem_spec
let result := implementation q w
use result
simp [result]
simp [implementation]
\end{lstlisting}
\end{minipage}
\begin{minipage}[t]{0.5\linewidth}
(d)
\begin{lstlisting}[basicstyle=\scriptsize\ttfamily]
def problem_spec
(implementation: List Int → Int → Bool)
(q: List Int) (w: Int) :=
let spec (result : Bool) :=
  (result → (List.Palindrome q)) ∧ 
  (result → (List.sum q ≤ w)) ∧
  (¬(List.Palindrome q) → ¬ result) ∧
  (¬(List.sum q ≤ w) → ¬ result)
∃ result, implementation q w = result ∧ spec result
\end{lstlisting}
(e)
\begin{lstlisting}[basicstyle=\scriptsize\ttfamily]
def implementation (q: List Int) (w: Int) : Bool :=
-- implementation generated by GPT-4o
let is_palindrome := q = q.reverse
let sum_le_w := q.sum ≤ w
is_palindrome && sum_le_w
\end{lstlisting}
(f)
\begin{lstlisting}[basicstyle=\scriptsize\ttfamily]
theorem correctness
(q: List Int) (w: Int)
: problem_spec implementation q w
:= by
-- proof generated by GPT-4o
unfold problem_spec
let result := implementation q w
use result
simp [result]
simp [implementation]
intro h -- <- The compilation fails here
simp [h]
exact List.eq_reverse_of_palindrome h.left
-- more proof trimmed
\end{lstlisting}
\end{minipage}
\end{mdframed}
%\reducevspacebetweenfigureandcaption
\vspace{-0.1in}
\caption{Illustration of specification leakage (left) and its mitigation (right) via non-computable specifications, using \humaneval problem 72. The task is to return \texttt{true} iff a list \texttt{q} is a palindrome and its sum is at most \texttt{w}. In (a–c), the spec is \emph{computable}: it encodes the desired logic in a Boolean expression, allowing the model to copy it directly in (b) and produce a trivial proof (c) via just unfolding and simplifying basic definitions used in the theorem statement. In contrast, (d–f) use a \emph{non-computable} spec expressed in \texttt{Prop} with logical implications. The corresponding implementation (e), generated by GPT-4o using few-shot prompting, reflects the semantic intent without mirroring the spec. The proof (f) fails without additional reasoning, highlighting the challenge of proving correctness when logic cannot be mechanically unfolded. Non-computable specs thus act as guardrails, requiring models to reason rather than copy.}
\label{fig:leak-issue-examples}
\vspace{-0.3in}
\end{figure}

Creating this benchmark involved substantial manual effort. On average, writing a formal specification took annotators \emph{25 minutes per problem} on average, with an additional 15 minutes spent reviewing each other’s specifications. Some problems involving complex non-computable specs required over an hour. To better understand problem difficulty and verify feasibility, we manually authored correctness proofs for a small random sample of benchmark problems. These ranged from \emph{10 lines (e.g., problem\_17) to 225 lines (e.g., problem\_0)}, reflecting a wide span of proof complexity.

In addition to the main benchmark, we \emph{release a small hand-curated few-shot prompt dataset} comprising of 5 problems distinct from \humaneval. All of these problems include hand-written implementations, and some of them additionally include manually written equivalence and isomorphism proofs. For example, one correctness proof spans 309 lines, while corresponding isomorphism proofs range from 29 to 82 lines. This auxiliary dataset is intended to support prompt tuning and evaluation in few-shot or in-context learning setups.

Curating the benchmark also revealed deeper challenges inherent to formal verification. For instance, in the \humaneval problem involving root-finding for polynomials (see \Cref{fig:poly-root-example}), proving termination is difficult due to reliance on unbounded numerical search. Similarly, generating verified code for “finding all prime Fibonacci numbers” encounters foundational roadblocks, as there is no known proof that infinitely many such numbers exist—highlighting how natural language tasks can conceal deep mathematical issues when formalized. One potential way to deal with these types of formulations is by adding the concept of computational fuel and approximate answers (see \Cref{fig:poly-root-example}, and \Cref{fig:hard-problem-examples} in \Cref{sec:hard-problem-examples}).
Writing \textit{non-computable} specifications is particularly challenging for problems that rely on language-level features like Python's \texttt{eval}, as seen in Problem 160. Since Lean lacks direct string-based evaluation, we had to reconstruct the behavior using inductive definitions over token lists and arithmetic expressions. This required converting a naturally computable task into a semantically equivalent, non-computable formulation without leaking implementation details. As shown in \Cref{fig:non-computable-example2} (in \Cref{sec:app-writing-non-computable-specification}), achieving this often involves layered recursive structures and careful abstraction to ensure both correctness and opacity.

\begin{figure}[t]
\centering
\vspace{-0.05in}
\begin{mdframed}[roundcorner=10pt, innerleftmargin=6pt, innerrightmargin=6pt, innertopmargin=4pt, innerbottommargin=4pt]
\begin{minipage}[t]{0.5\linewidth}
(a)
\begin{lstlisting}[basicstyle=\scriptsize\ttfamily]
def problem_spec
-- function signature
(implementation: List Rat → Rat)
-- inputs
(xs: List Rat) :=
-- spec
let spec (result: Rat) :=
  let eps := (1: Rat) / 1000000;
  xs.length ≥ 1 → xs.length % 2 = 0 →
  ∀ poly : Polynomial Rat,
    poly.degree = some (xs.length - 1) →
    (∀ i, i ≤ xs.length - 1 → poly.coeff i = xs.get! i) →
    |poly.eval result| ≤ eps;
-- program termination
∃ result,
  implementation xs = result ∧
  spec result
\end{lstlisting}
\end{minipage}
\hfill
\begin{minipage}[t]{0.5\linewidth}
(b)
\begin{lstlisting}[basicstyle=\scriptsize\ttfamily]
-- possible implementation using Newton's method
def implementation (xs: List Rat) : Rat :=
let rec poly (xs: List Rat) (x: Rat) := xs.reverse.foldl (λ acc a => acc * x + a) 0;
let rec poly' (xs: List Rat) (x: Rat) := (xs.drop 1).reverse.foldl (λ acc a => acc * x + a) 0;
let rec eps := (1: Rat) / 1000000;
let rec find_zero (xs: List Rat) (guess: Rat) (fuel: Nat) :=
let eval := poly xs guess;
let eval' := poly' xs guess;
if eval ≤ eps ∨ fuel = 0 then (guess, fuel)
else
let guess' := (eval' * guess - eval) / eval';
find_zero xs guess' (fuel - 1);
(find_zero xs 1.0 1000000).1
\end{lstlisting}
\end{minipage}

\end{mdframed}

\vspace{-0.1in}
\caption{\textbf{Polynomial Root-Finding.}
Problem 32 asks for an approximate real root of a degree-\(n\) polynomial. The spec enforces small residual error (\(<10^{-6}\)). The implementation uses Newton's method with bounded recursion; proving termination is non-trivial due to lack of guaranteed derivative behavior.}
\label{fig:poly-root-example}
\vspace{-0.2in}
\end{figure}

Another instructive case is the problem of computing the MD5 checksum (problem 162). Here, the formal specification must, by necessity, describe the exact computation, making it closely related to the implementation itself. Since we could not find any popular hashing libraries in Lean, we chose not to formalize this specific problem. However, we prescribe the recipe for creating non-computable definitions in \Cref{sec:app-writing-non-computable-specification}, given that we know the computable definition.

While adapting \humaneval to Lean, we encountered several language-level limitations. Some problems relying on dynamic typing or polymorphic return types—like Python’s \texttt{Any}—could not be faithfully represented in a statically typed setting (e.g., problems 22 and 137). As a result, we were able to formalize 161 out of the original 164 problems. In problem 103, where the output is either a binary string or \texttt{None} based on input validity, we use \texttt{Option String} as the return type. In problem 129, where the function may return either a list of words or a number, we encode this using disjoint union type in Lean: \texttt{(List String)} $\oplus$ \texttt{Nat}, allowing only one of the two values to be populated at a time.

Prior work, such as FVAPPS~\cite{dougherty2025provingcodinginterviewbenchmark}, relies on automatically generated specifications that can be incomplete or leaky, allowing trivial implementations (e.g., always returning zero) to pass (see \Cref{fig:fvapps-example} in \Cref{sec:fvapps}). Our human-curated specifications ensure completeness and robustness, closing such loopholes and surfacing the real verification complexity hidden in everyday programming problems.

% \stefan{Could also mention that we had to use Lean's Option type, because we can't return either a string or -1 as in Python. See e.g., problem 103}

% \section{\clever Prover: Approach}
% \input{clever_prover_approach}

\section{Evaluation}
We evaluated several state-of-the-art LLMs and agentic approaches on \clever. Now we elaborate on the results.

\paragraph{Evaluation Metric.}
To fairly compare approaches that differ in model size, latency, and API usage, we adopt the metric \textit{pass@k-seconds}—the fraction of benchmark problems solved within a fixed time budget \(k\). A task is marked as solved only if both the formal specification and the implementation are generated and verified via Lean’s type checker. As described in \Cref{fig:evaluation-strategy}, each step in the \clever pipeline (spec generation, equivalence proof, implementation, and correctness proof) is retried until a valid Lean-compilable output is found or the time runs out.

We also compare the different approaches for each task using \emph{pass@k}\cite{chen2021evaluating}  (\emph{compile@k} and \emph{prove@k} respectively). 

\begin{wrapfigure}{r}{0.6\textwidth}
\vspace{-0.25in}
{\footnotesize
\begin{codebox}
\li \Comment \textit{Assume \proc{Retry} retries the given function}
\li \Comment \textit{until it generates compilable Lean 4 code or timeouts.}
\li \Comment \textit{\proc{Retry} returns the Lean 4 code and remaining time.}
\Procname{$\proc{Evaluate}(approach,\ \text{timeout})$}
\li $t_{\text{rem}} \gets \text{timeout}$
\li $\SpecF,\ t_{\text{rem}} \gets$ 
    \proc{Retry}(\texttt{GenerateSpec}, $\SpecNL$, $t_{\text{rem}}$)
\li $P_{\text{eq}},\ t_{\text{rem}} \gets$ 
    \proc{Retry}(\texttt{ProveEquivalence}, $(\SpecF,\ \SpecF^*)$, $t_{\text{rem}}$)
\li \If $t_{\text{rem}} \leq 0$ \Return \textbf{Fail}
\li $\Program,\ t_{\text{rem}} \gets$ 
    \proc{Retry}(\texttt{GenerateImpl}, $(\SpecNL,\ \SpecF)$, $t_{\text{rem}}$)
\li $P_{\chi},\ t_{\text{rem}} \gets$ 
    \proc{Retry}(\texttt{ProveCorrectness}, $(\Program,\ \SpecF^*)$, $t_{\text{rem}}$)
\li \If $t_{\text{rem}} \leq 0$ \Return \textbf{Fail}
\li \Return \textbf{Success (all Lean 4 checks passed)}
\end{codebox}
}
\vspace{-0.2in}
\caption{Evaluation strategy: retry each generation step until Lean compilation succeeds or a timeout is reached.}
\label{fig:evaluation-strategy}
\vspace{-0.2in}
\end{wrapfigure}
%\end{figure}

\paragraph{Evaluated Baselines.}
% \amit{TODO: Add DeepSeek baseline too}
We evaluate three families of approaches for end-to-end verified code generation. The \textbf{Few-Shot Baseline} uses large language models (\texttt{GPT-4o}, \texttt{Claude-3.7}, \texttt{o4-mini}, and \texttt{DeepSeek-R1}) to generate all components—specifications, implementations, and proofs—via few-shot prompting with 1–2 exemplars. This baseline assesses the raw capability of LLMs to reason formally without task-specific training or tooling. The \textbf{COPRA Baseline} replaces the proof generation steps (Stages 2 and 4) with COPRA \cite{thakur2024incontext}, a neuro-symbolic proof search agent designed to produce Lean-compatible proofs when provided with an off-the-shelf foundational model and a Lean theorem statement to prove. This setup isolates proof search difficulty from the upstream generation task.

\begin{table}[t]
\centering
\scalebox{0.75}{
\begin{tabular}{@{}lllllccccc@{}}
\toprule
\multicolumn{4}{c}{\textbf{Approach Components}} & 
\multicolumn{5}{c}{\textbf{Pass@k-sec}} \\
& & & & &
\multicolumn{2}{c}{\textbf{Spec Cert.}} & 
\multicolumn{2}{c}{\textbf{Impl Cert.}} & 
\multicolumn{1}{c}{\textbf{End-to-End}} \\
\cmidrule(lr){1-5} \cmidrule(lr){6-7} \cmidrule(lr){8-9}
Model & Spec Gen & Equiv Proof & Impl Gen & Corr Proof & \textit{Compiled} & \textit{Proved} & \textit{Compiled} & \textit{Proved}  \\
\midrule
\multicolumn{5}{c}{Few-Shot Baseline} \\
% Few-Shot Baselines & & &  & & &  & & &  &  \\
\cmidrule(lr){1-5}
GPT-4o & FS & FS & FS & FS & 84.472\% & 0.621\% & 68.323\% & 0.621\% & 0\% \\
o4-mini & FS & FS & FS & FS & 82.609\% & 1.242\% & 83.230\% & 1.863\%  & 0.621\% \\
Claude-3.7 & FS & FS & FS & FS & 86.957\% & 0.621\% & 65.217\% & 1.863\%  & 0.621\% \\
DeepSeek-R1 & FS & FS & FS & FS & 71.42\% & 0.621\% & 60.870\% & 5.559\%  & 0.621\% \\
\midrule
\multicolumn{5}{c}{COPRA Baseline} \\
% Few-Shot Baselines & & &  & & &  & & &  &  \\
\cmidrule(lr){1-5}
GPT-4o & FS & COPRA & FS & COPRA & 76.398\% & 1.863\% & 68.323\% & 3.727\%  & 0.621\% \\
Claude-3.7 & FS & COPRA & FS & COPRA & 81.366\% & 1.242\% & 65.217\% & 8.696\% & 0.621\% \\
\midrule
\multicolumn{5}{c}{Hybrid Baseline} \\
\cmidrule(lr){1-5}
GPT-5-mini & FS & Kimina FS & FS & Kimina FS & 90.062\% & 0\% & 84.472\% & 0.621\% & 0\% \\
\bottomrule
\end{tabular}
}
\vspace{0.1in}
\caption{
Evaluation of different strategies for \textbf{end-to-end verified code generation}. Each approach consists of five components: \textbf{Model} (LLM used), \textbf{Spec Gen} (formal specification generation), \textbf{Equiv Proof} (proof of equivalence to ground-truth spec), \textbf{Impl Gen} (program synthesis), and \textbf{Corr Proof} (proof of implementation correctness). \textbf{FS} indicates few-shot prompting with 1--2 examples. Evaluation follows the pipeline in \Cref{fig:evaluation-strategy}. \textbf{Pass@k-seconds} with \(k = 600\) reports the fraction of tasks where Lean successfully compiles the outputs and accepts the associated proofs within a 600-second time budget. The \textbf{Compiled} columns indicate whether the generated Lean code is syntactically valid and type-checks. The \textbf{Proved} columns reflect whether the corresponding proofs were accepted by Lean's kernel, thereby certifying semantic correctness. The \textbf{End-to-End} column reports full pipeline success---i.e., both the specification and implementation must compile and their respective proofs must be accepted. Despite strong models like GPT-4o achieving high compilation rates, formal correctness remains challenging: no approach has yet succeeded across all stages on more than one problem (specifically problem 53).
}
\label{tab:evaluation-detailed}
\vspace{-0.35in}
\end{table}

\paragraph{Results.}
Our primary evaluation metric focuses strictly on semantic correctness: a task is considered successful only if both the specification and the implementation are formally certified via Lean proofs. This strict definition ensures that reported scores reflect genuine end-to-end verification. However, to better diagnose failure modes, we also report auxiliary statistics: the fraction of tasks where generated specifications and implementations \emph{compile} successfully. These serve as proxies for the model’s fluency in Lean and its ability to produce well-typed artifacts.

\begin{figure}[t!]
    \centering
    \small % Keeps captions and axis labels smaller
    
    % Row 1: Specifications
    \begin{subfigure}[b]{0.48\textwidth}
        \includegraphics[width=0.9\linewidth]{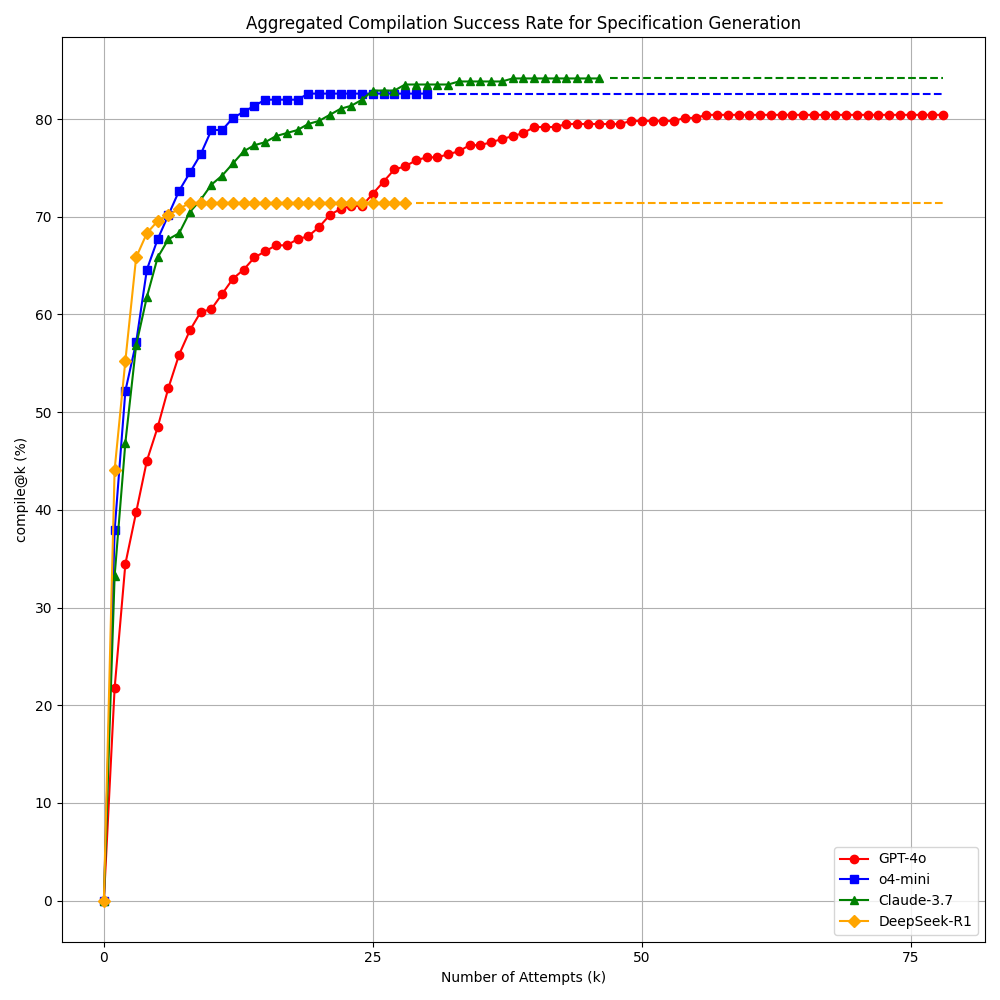} % <-- REDUCED IMAGE WIDTH
        \caption{Specification Compilation (\texttt{compile@k})}
        \label{fig:spec-compile}
    \end{subfigure}
    \hfill
    \begin{subfigure}[b]{0.48\textwidth}
        \includegraphics[width=0.9\linewidth]{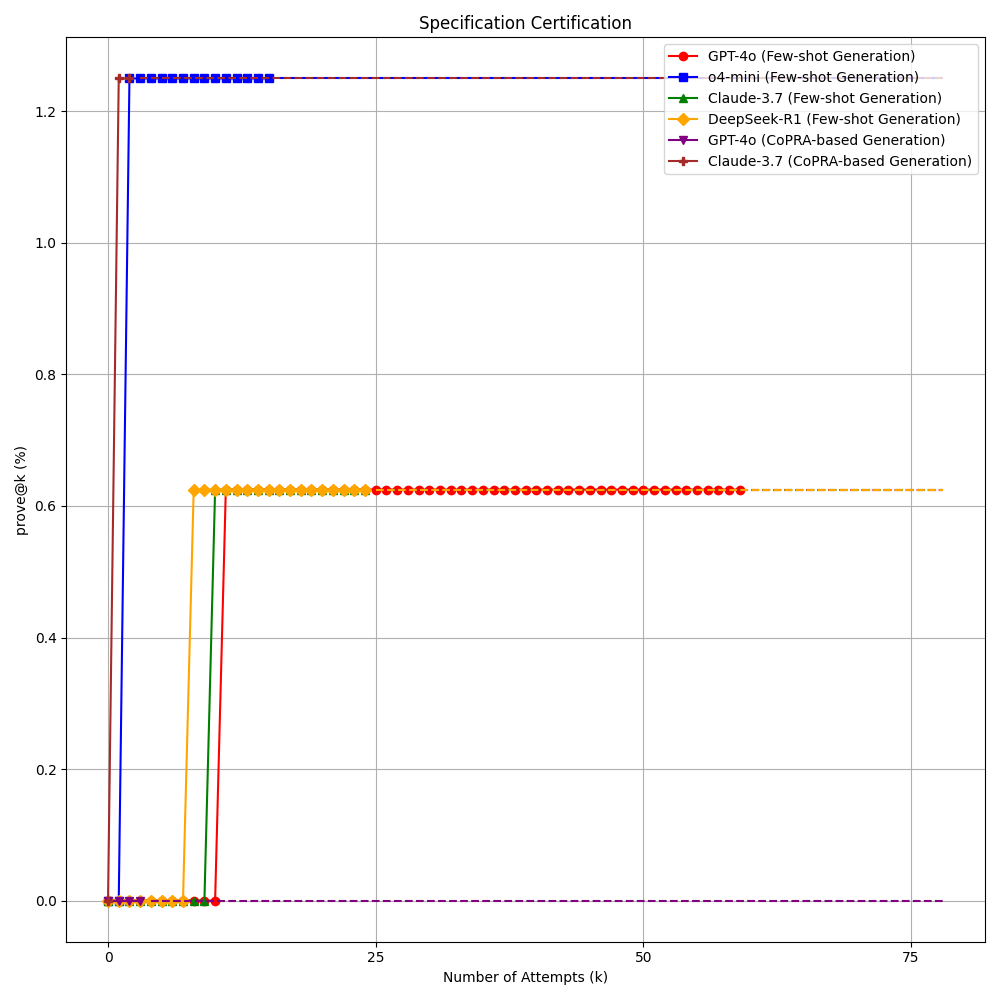} % <-- REDUCED IMAGE WIDTH
        \caption{Specification Certification (\texttt{prove@k})}
        \label{fig:spec-prove}
    \end{subfigure}
    
    \vspace{0.05cm} % Even tighter vertical space between rows
    
    % Row 2: Implementations
    \begin{subfigure}[b]{0.48\textwidth}
        \includegraphics[width=0.9\linewidth]{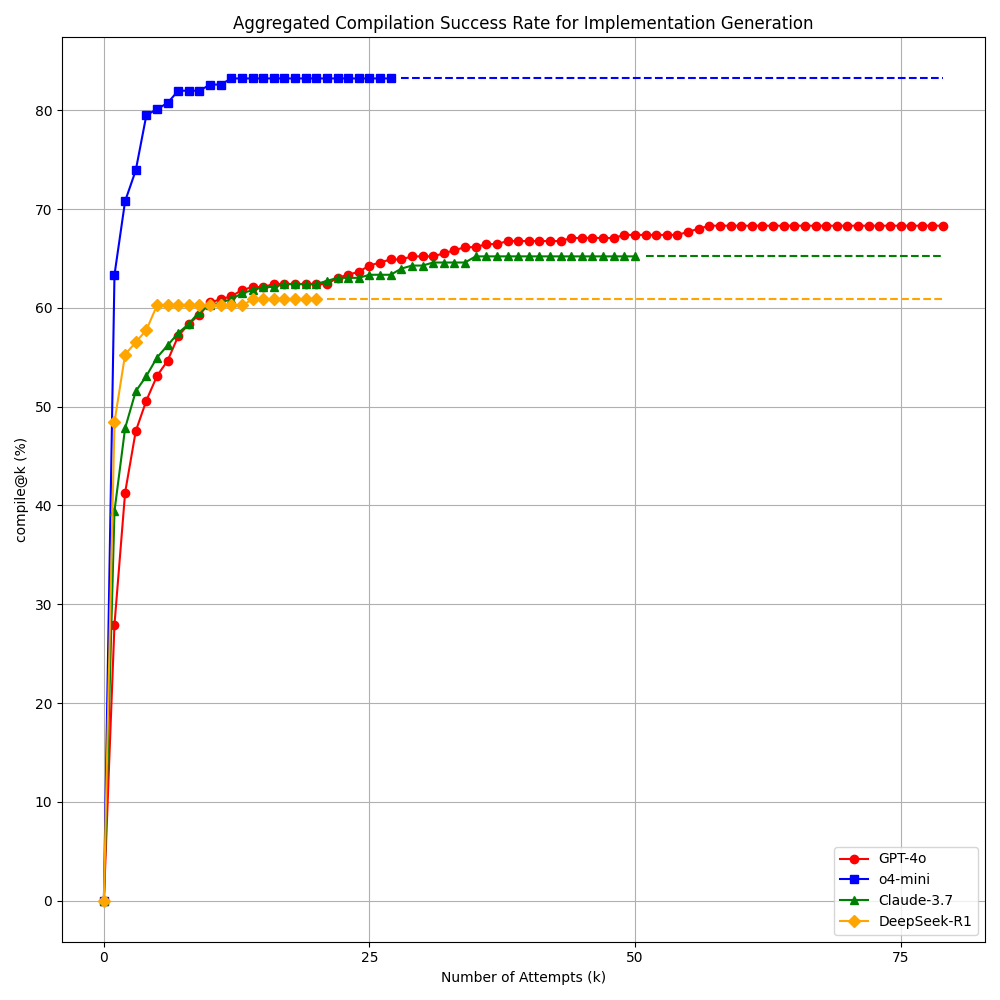} % <-- REDUCED IMAGE WIDTH
        \caption{Implementation Compilation (\texttt{compile@k})}
        \label{fig:impl-compile}
    \end{subfigure}
    \hfill
    \begin{subfigure}[b]{0.48\textwidth}
        \includegraphics[width=0.9\linewidth]{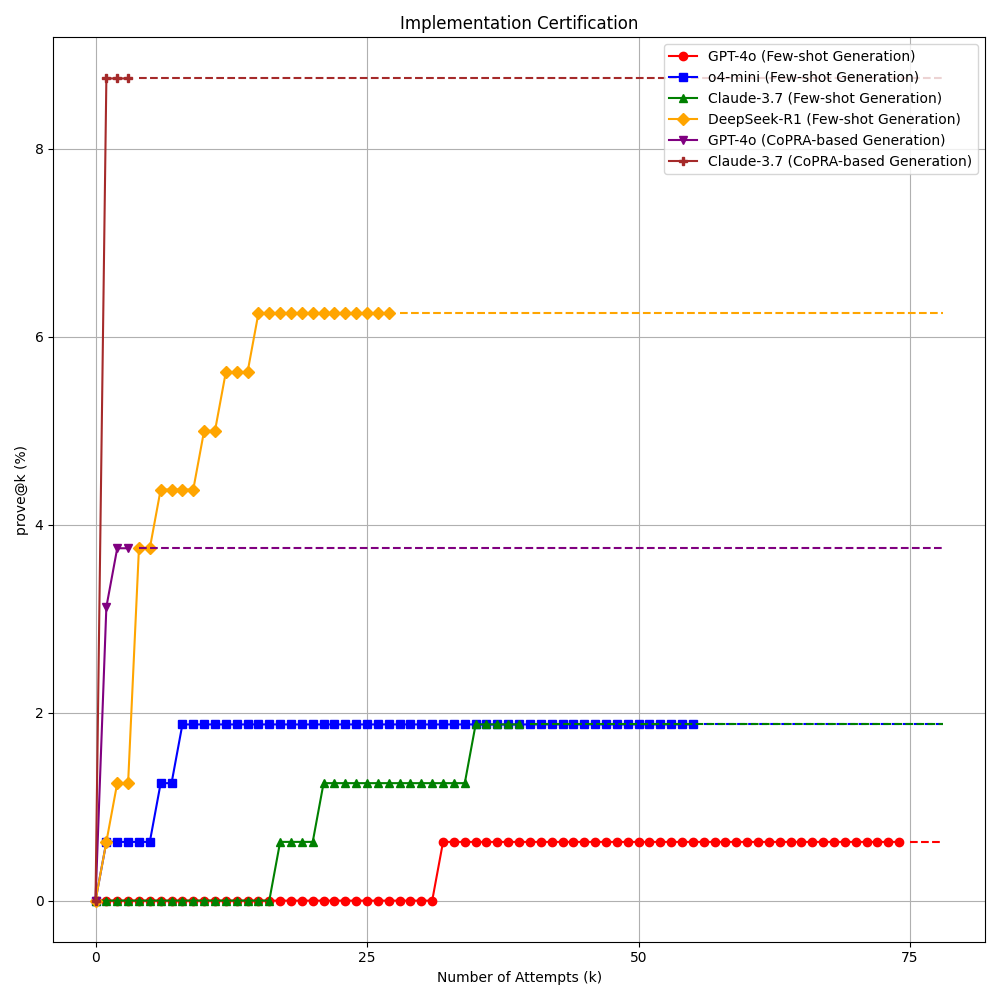} % <-- REDUCED IMAGE WIDTH
        \caption{Implementation Certification (\texttt{prove@k})}
        \label{fig:impl-prove}
    \end{subfigure}

    \caption{
    Aggregated \texttt{compile@k} and \texttt{prove@k} results across $k$ attempts, diagnosing failure modes for specification and implementation generation. Dotted lines indicate extrapolation of the pass rate beyond the 600s timeout.
        \textbf{(a) Specification Compilation:} Most models achieve high \texttt{compile@k} rates, with \texttt{Claude-3.7} and \texttt{o4-mini} reaching >80\%.
        \textbf{(b) Specification Certification:} Proving specification equivalence is a major bottleneck. All few-shot models solve only one problem (0.62\%), while \texttt{GPT-4o (CoPRA)} proves ~1.8\%.
        \textbf{(c) Implementation Compilation:} \texttt{o4-mini} has the highest compilation rate at >80\%, while other models cluster between 60-70\%.
        \textbf{(d) Implementation Certification:} \texttt{Claude-3.7 (CoPRA)} performs best, certifying 8.7\% of implementations, followed by \texttt{DeepSeek-R1} (Few-shot) at 5.6\%.
        These plots show that while models are fluent at generating \emph{compilable} artifacts, formal \emph{certification} remains the key challenge, with all \texttt{prove@k} rates below 10\%.
    }
    \label{fig:pass-at-k-results}
    %\vspace{-0.45in} % Increased negative vertical space at the end
\end{figure}

In particular, implementation compilation includes not only type-checking against the declared function signature, but also validation against a suite of example-based test cases adapted from the original \humaneval prompts. While passing these tests provides some evidence of functional correctness\cite{evalplus}, we deliberately exclude them from our core success metric—since test cases offer only partial coverage and cannot guarantee semantic soundness (see \Cref{sec:problem-formulation} for discussion).
% \stefan{This seems like a good place to reference HumanEval+ (https://github.com/evalplus/evalplus)}
%\amit{Mentioned}

%As shown in \Cref{tab:evaluation-detailed}, compilation rates are broadly similar across few-shot models for both specification and implementation generation. A notable exception is the higher implementation compilation rate achieved by \texttt{o4-mini}, which contrasts with its lower success in proving correctness. More generally, even when an approach successfully certifies multiple specifications or verifies correctness for multiple implementations, the overall end-to-end success rate remains low. This is largely due to \emph{mismatch}: tasks for which specification certification is tractable are often those where implementation correctness proofs are especially difficult, and vice versa. As a result, the joint success condition is rarely satisfied.  

%[This is NEW]
As shown in \Cref{fig:pass-at-k-results} and \Cref{tab:evaluation-detailed}, compilation rates are broadly similar across few-shot models for both specification (\Cref{fig:spec-compile}) and implementation (\Cref{fig:impl-compile}) generation. A notable exception is the higher implementation compilation rate achieved by \texttt{o4-mini}, which contrasts with its lower success in proving correctness. More generally, even when an approach successfully certifies multiple specifications (\Cref{fig:spec-prove}) or verifies correctness for multiple implementations (\Cref{fig:impl-prove}), the overall end-to-end success rate remains low. This is largely due to \emph{mismatch}: tasks for which specification certification is tractable are often those where implementation correctness proofs are especially difficult, and vice versa. As a result, the joint success condition is rarely satisfied.

Another interesting observation is that \texttt{Claude-3.7}, when used along with COPRA, can certify more implementations (14) than all other models; however, its performance on specification certification is only comparable to other models. We believe that this might have to do with the length of proofs needed for specification certification, and hence, in the limited timeout it is hard to find the full proof for specification.

Finally, we evaluated \emph{KiminaProver-7b}\cite{wang2025kimina}, a specialized
prover. Due to its non-standard output formatting which posed parsing challenges,
we used it only for proof steps, pairing it with \texttt{GPT-5-mini} for
generation. This hybrid (\Cref{tab:evaluation-detailed}) yielded the
\textbf{highest compilation rates} (90.1\% spec, 84.5\% impl), highlighting
\texttt{GPT-5-mini}'s fluency. However, it certified almost nothing (0 spec, 1 impl),
as the pipeline was bottlenecked by these same parsing issues.

\paragraph{Proof Difficulty and Structure.}
As shown in \Cref{tab:qualitative-analysis}, proofs for \textbf{specification certification}
% \stefan{In sections 1/2, "specification/implementation certification" is capitalized}
are consistently longer and harder to generate than those for implementation correctness. This is expected: proving that a generated spec is semantically equivalent to a non-computable reference specification requires models (or agents) to reason abstractly about intent, without access to implementation-level cues. In contrast, correctness proofs often benefit from direct pattern matching or automation through tactics like \texttt{simp}.

This distinction is especially evident in the only problem for which an end-to-end verified code generation succeeds across multiple models: \textbf{problem 53}, which asks for the sum of two integers. Despite the simplicity of the implementation, the ground-truth specification is expressed in a way that deliberately obfuscates the target behavior. This design makes the equivalence proof non-trivial and requires models (or COPRA) to recover the algebraic structure underlying addition. Even here, success is only possible because the proofs admit aggressive automation via \texttt{simp} and \texttt{ring}. The full problem is shown in \Cref{fig:add-example}, which illustrates the separation between syntactic and semantic difficulty across spec, implementation, and proofs. 

Notably, \texttt{Claude-3.7} in combination with COPRA successfully solves every implementation certification task that any other approach is able to solve. \Cref{fig:brazilian-factorial-example} in \Cref{sec:app-proofs-found} illustrates one such case, showcasing a 35-line proof for the Brazilian factorial task that requires symbolic reasoning over factorial identities and recursive structure.

Unlike math-focused benchmarks such as MiniF2F \cite{zheng2021minif2f}, where many proofs are short, goal-directed, and amenable to automation via tactics like \texttt{linarith}, \texttt{ring}, or \texttt{simp}, the proofs in our benchmark often mirror the \emph{control flow} and \emph{branching structure} of programs. As a result, standard automation is rarely sufficient. Correctness proofs frequently require reasoning case-by-case over pattern-matched inputs, recursive call structure, or multiple conditional branches. Even when the final goal involves simple arithmetic, the surrounding structure demands explicit handling of recursive unrolling, constructor cases, or fuel-based invariants. For example, proving correctness for recursive implementations like factorial products or root-finding procedures involves handling termination branches, intermediate values, and variable dependencies that make tactics like \texttt{linarith} or \texttt{ring} ineffective without significant manual decomposition. This structurally rich proof landscape contrasts with the often-flat logical forms seen in MiniF2F and underscores the need for symbolic agents like COPRA that can perform guided proof search beyond tactic chaining. 

% \jasper{Table 2 caption says ``qualitative analysis" but does it still count as that if all the data is numerical?}

\begin{table}[t]
\centering
\scalebox{0.9}{
\begin{tabular}{@{}llcccccc@{}}
\toprule
\textbf{Model} & \textbf{Approach} & \textbf{Certification} & \textbf{\# Qed} & \textbf{Avg. \# Lines} & \textbf{\# Line (Min-Max)} & \textbf{Avg. Time (s)}\\
\midrule
GPT-4o & FS & Spec & 1 & 16.0 & 16–16 & 124.3\\
GPT-4o & FS & Impl & 1 & 6.0 & 6–6 & 291.6 \\
o4-mini & FS & Spec & 2 & 29.5 & 26–33 & 87.0\\
o4-mini & FS & Impl & 3 & 14.0 & 10–21 & 204.0\\
Claude-3.7 & FS & Spec & 1 & 38.0 & 38–38 & 195.7\\
Claude-3.7 & FS & Impl & 3 & 12.7 & 6–21 & 414.4\\
DeepSeek-R1 & FS & Spec & 1 & 26.0 & 26–26 & 170.8\\
DeepSeek-R1 & FS & Impl & 9 & 14.1 & 3–27 & 137.73\\
\midrule
GPT-4o & COPRA & Spec & 3 & 26.3 & 16–44 & 97.9 \\
GPT-4o & COPRA & Impl & 6 & 10.8 & 6-19 & 199.6 \\
Claude-3.7 & COPRA & Spec & 2 & 30.5 & 16-45 & 308.7 \\
Claude-3.7 & COPRA & Impl & 14 & 14.3 & 4–35 & 165.8 \\
\bottomrule
\end{tabular}
}
\caption{Analysis of successfully generated proofs across different models and certification types. We report: (1) the number of problems for which the correctness (isomorphism resp.) proofs are found by the approach in the column ``\# Qed'' (see \Cref{fig:clever-summary}), (2) the average number of lines in the proof, (3) the range of proof lengths (min–max), and (4) the average time it took for the approach to find a proof (given a proof was found). This analysis highlights variation in proof complexity and model behavior across settings. Few-shot prompting typically yields shorter, more brittle proofs, while COPRA-augmented configurations show higher robustness, with more consistent success and a broader range of proof strategies. Proof line counts serve as a coarse indicator of reasoning complexity.}
\label{tab:qualitative-analysis}
\vspace{-0.2in}
\end{table}

\begin{figure}[t]
\centering
\vspace{-0.05in}
\begin{mdframed}[roundcorner=6pt, innerleftmargin=6pt, innerrightmargin=6pt, innertopmargin=4pt, innerbottommargin=4pt]

\begin{minipage}[t]{0.49\linewidth}
(a)
\begin{lstlisting}[basicstyle=\scriptsize\ttfamily]
def problem_spec (impl : Int → Int → Int) (x y : Int) :=
  let spec (res : Int) := res - x - y = 0
  ∃ result, impl x y = result ∧ spec result
\end{lstlisting}
(b)
\begin{lstlisting}[basicstyle=\scriptsize\ttfamily]
def generated_spec (impl : Int → Int → Int) (x y : Int) : Prop :=
  impl x y = x + y
\end{lstlisting}
(c)
\begin{lstlisting}[basicstyle=\scriptsize\ttfamily]
def implementation (x y : Int) : Int := x + y
\end{lstlisting}
% \end{minipage}
% \hfill
% \begin{minipage}[t]{0.49\linewidth}
(d)
\begin{lstlisting}[basicstyle=\scriptsize\ttfamily]
theorem correctness (x y : Int) : problem_spec implementation x y :=
by
  unfold problem_spec
  let result := implementation x y
  use result
  simp [result]
  simp [implementation]
\end{lstlisting}
\end{minipage}
\hfill
\begin{minipage}[t]{0.49\linewidth}
(e)
\begin{lstlisting}[basicstyle=\scriptsize\ttfamily]
theorem spec_isomorphism :
  ∀ impl, (∀ x y, problem_spec impl x y) ↔
           (∀ x y, generated_spec impl x y) :=
by
  intro impl
  apply Iff.intro
  -- → direction
  intro h_prob_spec
  intro x y
  have h := h_prob_spec x y
  simp [generated_spec, problem_spec] at h
  rw [generated_spec]
  linarith
  -- ← direction
  intro h_gen_spec
  intro x y
  unfold problem_spec
  simp
  have h := h_gen_spec x y
  simp [generated_spec] at h
  rw [h]
  ring
\end{lstlisting}
\end{minipage}

\end{mdframed}

\vspace{-0.1in}
\caption{
\textbf{End-to-end verified example: Problem 53 (Add Two Numbers).}
This task requires adding two integers \texttt{x} and \texttt{y}. Shown are all components of the certification pipeline: (a) a non-computable ground truth spec using subtraction to hide the implementation, (b) the model-generated spec, (c) the implementation \texttt{x + y}, (d) a short correctness proof, and (e) an isomorphism proof relating the two specs. While the implementation is simple, the spec equivalence proof requires symbolic reasoning. This is the only HumanEval-derived task with full verification across multiple approaches.
}
% \caption{
% \textbf{End-to-end verified example: Problem 53 (Add Two Numbers).}
% This problem asks the model to implement a function that returns the sum of two integers \texttt{x} and \texttt{y}. Shown here are all components needed to pass the full certification pipeline: (a) the \textbf{ground truth specification} written in a non-computable form using the fact that subtraction is an inverse operation for addition (without leaking the actual implementation directly) (b) the \textbf{model-generated specification}, which defines the behavior directly; (c) the \textbf{implementation} \texttt{x + y}; (d) the \textbf{correctness proof} showing that the implementation satisfies the reference specification; and (e) the \textbf{specification equivalence proof}, which establishes that the generated spec matches the hidden reference. While the implementation is trivial, the isomorphism proof requires symbolic reasoning to relate semantically equivalent specifications. In contrast, the correctness proof is short and automatable using tactics like \texttt{simp} and \texttt{ring}. This is the only HumanEval-derived task for which full verification was achieved across multiple approaches.
% }
\label{fig:add-example}
\vspace{-0.2in}
\end{figure}

\section{Related Work}
\textbf{Formal Verification.} Formal verification encompasses a range of techniques aimed at mathematically proving the correctness of software or hardware systems with respect to a formal specification, thereby providing strong guarantees beyond traditional testing. Dafny and Verus \citep{dafny, verus} utilize SMT solvers to perform verification given proper verification conditions. Interactive theorem provers like Lean, Isabelle, and Coq \citep{de2015lean, paulson1994isabelle, huet1997coq} offer highly expressive logics where users construct proofs interactively with tactic-based automation. Notably, interactive theorem provers have been involved in the verification of C compilers, microkernels, and distributed systems protocols \citep{compcert, sel4, verdi}.

\textbf{Benchmarks.}
Recent efforts have developed benchmarks for formal verification with the onset of powerful neural models. FVAPPS \citep{dougherty2025provingcodinginterviewbenchmark} uses an LLM on scraped competition problems to automatically create formal specifications for 4715 problems, 1083 of which are guarded with test cases. However, the formal specifications themselves are often easily hackable (see \Cref{sec:fvapps}), with verification correctness guarded by a layer of test cases. Here, we aim to provide complete formal specifications, which cannot be done accurately with automatic annotation. miniCodeProps \citep{lohn2024minicodepropsminimalbenchmarkproving} contains 201 verification problems regarding data structures and induction problems; however, they do not include specification synthesis or equivalence tests. DafnyBench \citep{loughridge2024dafnybenchbenchmarkformalsoftware} is a benchmark of 782 stand-alone Dafny programs collected from prior benchmarks and Dafny repositories, where the synthesis task is to generate the verification conditions that allow Dafny to prove correctness. At the time, the best model was \texttt{Claude 3 Opus} which solved $\approx$ 68 \% of the problems. Software engineering benchmarks have become extremely popular in recent literature, including benchmarking performance fixing real-world issues \citep{jimenez2024swebenchlanguagemodelsresolve} and contamination-free code generation \citep{jain2024livecodebenchholisticcontaminationfree}. In our work, we employ \humaneval \citep{chen2021evaluating} to create \textsc{Clever}, our formal verification and synthesis benchmark.
Formal verification is also applied in mathematical domains. Mathlib \citep{mathlib} and the Archive of Formal Proofs \citep{ArchiveFormalProofs} constitute formal mathematical repositories in Lean and Isabelle respectively, from which benchmarks have been derived \citep{hu2025minictxneuraltheoremproving, jiang2021lisa}. ProofNet \citep{azerbayev2023proofnetautoformalizingformallyproving} serves as a benchmark for producing proper specifications of mathematical problems. PutnamBench \citep{tsoukalas2024putnambenchevaluatingneuraltheoremprovers} is a formal benchmark of undergraduate-level competition problems in Lean, Isabelle, and Coq.
% \stefan{More papers: 
% https://arxiv.org/abs/2405.01787, 
% https://arxiv.org/abs/2501.16207v2,
% https://arxiv.org/abs/2503.14183,
% https://arxiv.org/abs/2310.17807, https://arxiv.org/abs/2402.00247}
% \amit{No space right now}

\textbf{Proving Methods.} Recent advances in neural models and LLMs have led to increased attention on formal verification and theorem-proving.  AlphaVerus \citep{aggarwal2024alphaverusbootstrappingformallyverified} introduces a tree search and refinement algorithm to self-improve at producing formally verified Verus code. Similarly, SAFE \citep{chen2024automatedproofgenerationrust} performs expert iteration in producing high-quality specification and proofs for generating verified Verus code. FVEL \citep{lin2024fvelinteractiveformalverification} uses symbolic methods to convert C programs into Isabelle, and then uses an LLM to generate correctness specifications which it then tries to prove. However, the automatic nature of the specification generation means correctness is not guaranteed. For mathematical theorem-proving, approaches involve tree search \citep{polu2020generative, yang2023leandojo}, reinforcement learning \citep{lin2025goedelproverfrontiermodelopensource, lample2022hypertree}, LLMs \citep{thakur2024incontext, xin2024deepseekprover}, and data augmentation and scale \citep{dong2025stpselfplayllmtheorem, deepmind2024alphaproof}.

\section{Conclusion}
We introduced a new benchmark for end-to-end verified code generation that shifts the focus from surface-level correctness to formal semantic guarantees. Unlike prior benchmarks that rely on test cases or computable specifications, our tasks are grounded in \emph{non-computable}, logic-based specifications that are explicitly designed to prevent implementation leakage. By enforcing a separation between specification intent and implementation behavior, the benchmark demands genuine reasoning rather than pattern matching or memorization.

Our evaluation protocol is deliberately staged, decomposing the pipeline into independently checkable phases: specification generation, isomorphism proof, implementation synthesis, and correctness proof. This staged design enables fine-grained diagnosis of where models succeed and fail—whether in interpreting informal intent, aligning it with formal meaning, or synthesizing verifiably correct programs. In particular, verifying the generated specification via \emph{isomorphism proofs} ensures semantic fidelity and introduces a novel opportunity: verified \emph{mining} of natural language and formal specification pairs from model generations, which could be reused for bootstrapping new training data.

Our benchmark introduces challenges beyond those in mathematical theorem-proving settings like miniF2F, where proofs are often short and tactic-friendly. In contrast, our tasks reflect the branching structure of real-world programs, requiring symbolic reasoning over control flow, recursion, and invariants—scenarios where automation alone breaks down. By combining structural complexity with formal soundness, non-leakage by design, and staged verification, the benchmark offers a rigorous, semantics-grounded testbed for verified code generation. It sets a new standard for advancing neural-symbolic reasoning toward scalable, trustworthy software verification.

\begin{ack}
This work was supported by NSF awards CCF-2212559 and CCF-2403211, and 
a 2025 Renaissance Philanthropy AI for Math award.
\end{ack}

%%%%%%%%%%%%%%%%%%%%%%%%%%%%%%%%%%%%%%%%%%%%%%%%%%%%%%%%%%%%

\newpage
\bibliography{references}
\bibliographystyle{references}

\newpage
\appendix
\section{Appendix}
\subsection{FVAPPS Benchmark}
\label{sec:fvapps}
% the formal specifications in these benchmarks tend not to capture the full (natural-language) intent behind the target program and sometimes hint at ways to implement the program (\Cref{app:fvapps})
The FVAPPS benchmark \citep{dougherty2025provingcodinginterviewbenchmark} is another code generation benchmark in Lean. However, unlike \clever, which requires a comprehensive proof of full program behavior, FVAPPS only requires the proof of a limited selection of properties of the program. The limitations of this are illustrated by the FVAPPS example in \Cref{fig:fvapps-example}. Here, a problem with a relatively complex natural language description only requires verifying lower-bound and upper-bound properties of the program implementation, as well as a few simple base cases. As can be seen, these properties are provably satisfied by a trivial program that always outputs 0 regardless of the input. Thus, it is clear that only requiring the proof of a small handful of properties does not capture the full intent of the natural language problem. This highlights the necessity of a verified code generation benchmark to require proofs of full program behavior, not just program properties.

\begin{figure}
    \centering
    \begin{mdframed}[roundcorner=10pt]
    \begin{lstlisting}
/--
solve_elections:
There are n voters, and two ways to convince each of them to vote for you. The first way to convince the $i$-th voter is to pay him $p_i$ coins. The second way is to make $m_i$ other voters vote for you, and the $i$-th voter will vote for free. Moreover, the process of such voting takes place in several steps. For example, if there are five voters with $m_1 = 1$, $m_2 = 2$, $m_3 = 2$, $m_4 = 4$, $m_5 = 5$, then you can buy the vote of the fifth voter, and eventually everyone will vote for you. Set of people voting for you will change as follows: $5$ → $1, 5$ → $1, 2, 3, 5$ → $1, 2, 3, 4, 5$. Calculate the minimum number of coins you have to spend so that everyone votes for you.
-/

def solve_elections (n : Nat) (voters : List (Nat × Nat)) : Nat := 0

theorem solve_elections_nonnegative (n : Nat) (voters : List (Nat × Nat)) : solve_elections n voters >= 0 :=
by rfl

theorem solve_elections_upper_bound (n : Nat) (voters : List (Nat × Nat)) : solve_elections n voters <= List.foldl (λ acc (pair : Nat × Nat) => acc + pair.2) 0 voters :=
Nat.zero_le _

theorem solve_elections_zero_votes (n : Nat) (voters : List (Nat × Nat)) : (List.all voters (fun pair => pair.1 = 0)) -> solve_elections n voters = 0 :=
fun _ => rfl

theorem solve_elections_single_zero_vote : solve_elections 1 [(0, 5)] = 0 :=
by rfl
    \end{lstlisting}
    \end{mdframed}
    \vspace{-0.1in}
    \caption{FVAPPS sample 23 and a trivial program that solves it, illustrating the limitations of not verifying full program behavior.}
    \label{fig:fvapps-example}
\end{figure}

\subsection{Hard to write Specifications}
\label{sec:hard-problem-examples}

\Cref{fig:hard-problem-examples} shows some problems for which the formal specification or the implementation is hard to write.

\begin{figure}
    \centering
    %\vspace{-0.1in}
    \begin{mdframed}[roundcorner=10pt]
    \begin{minipage}[t]{0.5\linewidth}
    (a)
    \begin{lstlisting}
def problem_spec
-- function signature
(implementation: List Rat → Rat)
-- inputs
(xs: List Rat) :=
-- spec
let spec (result: Rat) :=
  let eps := (1: Rat) / 1000000;
  xs.length ≥ 1 → xs.length % 2 = 0 →
  ∀ poly : Polynomial Rat,
    poly.degree = some (xs.length - 1) →
    (∀ i, i ≤ xs.length - 1 → poly.coeff i = xs.get! i) →
    |poly.eval result| ≤ eps;
-- program termination
∃ result,
  implementation xs = result ∧
  spec result

-- possible implementation using Newton's method
def implementation (xs: List Rat) : Rat :=
let rec poly (xs: List Rat) (x: Rat) := xs.reverse.foldl (λ acc a => acc * x + a) 0;
let rec poly' (xs: List Rat) (x: Rat) := (xs.drop 1).reverse.foldl (λ acc a => acc * x + a) 0;
let rec eps := (1: Rat) / 1000000;
let rec find_zero (xs: List Rat) (guess: Rat) (fuel: Nat) :=
let eval := poly xs guess;
let eval' := poly' xs guess;
if eval ≤ eps ∨ fuel = 0 then (guess, fuel)
else
let guess' := (eval' * guess - eval) / eval';
find_zero xs guess' (fuel - 1);
(find_zero xs 1.0 1000000).1
    \end{lstlisting}
    \end{minipage}
    % \vspace{-7.3in}
    % (b)\vspace{7.3in}
    \begin{minipage}[t]{0.5\linewidth}
    (b)
    \begin{lstlisting}
def problem_spec
-- function signature
(implementation: Nat → Nat)
-- inputs
(n: Nat) :=
-- spec
let spec (result: Nat) :=
  n > 0 →
    (∃ i, Nat.fib i = result ∧ Nat.Prime result ∧
      (∃! S : Finset Nat, S.card = n - 1 ∧
      (∀ y ∈ S, (∃ k, y = Nat.fib k) ∧ y < result ∧ Nat.Prime y))
    )

-- implementation without proof of
-- termination
def implementation (n: Nat) : Nat :=
let rec fib_prime (n: Nat) (i: Nat) : Nat :=
  if Nat.Prime (Nat.fib i) then
    if n = 1 ∨ n = 0
    then Nat.fib i
    else fib_prime (n - 1) (i + 1)
  else fib_prime n (i + 1)
termination_by n
decreasing_by
  -- Proof of termination is open problem
  sorry
  sorry
fib_prime n 0
    \end{lstlisting}
    \end{minipage}
    \end{mdframed}
    %\reducevspacebetweenfigureandcaption
    \vspace{-0.1in}
    \caption{
    Examples of benchmark challenges. (a) Polynomial root-finding: difficulties in proving termination of numerical search; (b) Prime Fibonacci finder: problem complexity rooted in the lack of a known proof of infinitude.}
    \label{fig:hard-problem-examples}
\end{figure}

\subsection{Writing non-computable specifications}
\label{sec:app-writing-non-computable-specification}

\Cref{fig:non-computable-example1} shows a computable vs non-computable version of the specification for finding the $n^{th}$ Fibonacci number. It can be observed that the computable version of the specification \textit{leaks} the implementation in contrast to the non-computable version. The non-computable specification uses an \textbf{inductive} definition of a recursive function.

\begin{figure}
    \centering
    %\vspace{-0.1in}
    \begin{mdframed}[roundcorner=10pt]
    \begin{minipage}[t]{0.5\linewidth}
    (a)
    \begin{lstlisting}
-- computable spec
def problem_spec
-- function signature
(implementation: List Nat → Nat)
-- inputs
(n: Nat) :=
-- spec
let spec (result: Nat) :=
  (n = 0 → result = 0) ∨
  (n = 1 → result = 1) ∨
  (2 ≤ n → ∃ fib_array : List Nat, 
   fib_array.length = n + 1 ∧
   fib_array[0]! = 0 ∧
   fib_array[1]! = 1 ∧
   (∀ i, 1 < i → i < n + 1 →
   fib_array[i]! = fib_array[i - 1]! + 
   fib_array[i - 2]!) ∧
   result = fib_array[n]!)
-- program termination
∃ result,
  implementation xs = result ∧
  spec result
    \end{lstlisting}
    \end{minipage}
    % \vspace{-7.3in}
    % (b)\vspace{7.3in}
    \begin{minipage}[t]{0.5\linewidth}
    (b)
    \begin{lstlisting}
-- non-computable spec
inductive fibonacci_non_computable : ℕ → ℕ → Prop
| base0 : fibonacci_non_computable 0 0
| base1 : fibonacci_non_computable 1 1
| step  : ∀ n f₁ f₂, 
fibonacci_non_computable n f₁ →
fibonacci_non_computable (n + 1) f₂ →
fibonacci_non_computable (n + 2) (f₁ + f₂)

def problem_spec
-- function signature
(implementation: Nat → Nat)
-- inputs
(n: Nat) :=
-- spec
let spec (result: Nat) :=
  fibonacci_non_computable n result
-- program termination
∃ result,
  implementation xs = result ∧
  spec result
    \end{lstlisting}
    \end{minipage}
    \end{mdframed}
    %\reducevspacebetweenfigureandcaption
    \vspace{-0.1in}
    \caption{
    Two different specs for finding the $n^{th}$ Fibonacci number. (a) shows a computable specification that \textit{leaks} the implementation; (b) shows a non-computable specification leading to no-leakage of the implementation and enforcing the model to learn the deeper logical inference.}
    \label{fig:non-computable-example1}
\end{figure}

Writing \textit{non-computable} specifications is a non-trivial task that requires a deep understanding of the problem. \Cref{fig:non-computable-example2} (problem 160) presents another complex example illustrating the difficulty of formulating such specifications. \Cref{fig:non-computable-example2} shows two versions of a specification for evaluating an expression given as a list of strings (\texttt{["2","+","3","*","4","-","5"]}). \Cref{fig:non-computable-example2}(a) evaluates the expression and later checks if the output matches the result (not specified in the figure), which is computable. \Cref{fig:non-computable-example2}(b) shows a non-computable version of the specification that checks if the result matches the output of evaluating the expression without leaking the implementation. One can notice that we need multiple inductive recursive definitions to ensure that the specification is clean and non-computable.

\begin{figure}
    \centering
    %\vspace{-0.1in}
    \begin{mdframed}[roundcorner=10pt]
    \begin{minipage}[t]{0.5\linewidth}
    (a)
    \begin{lstlisting}
inductive Op where
  | add | sub | mul | floordiv
deriving Repr, DecidableEq

def parseOp : String → Option Op
  | "+" => some .add | "-" => some .sub
  | "*" => some .mul | "//" => some .floordiv
  | _ => none

def precedence : Op → Nat
  | .mul | .floordiv => 2
  | .add | .sub      => 1

def apply : Op → Int → Int → Int
  | .add, a, b => a + b
  | .sub, a, b => a - b
  | .mul, a, b => a * b
  | .floordiv, a, b => a / b

inductive Tok where
  | num : Int → Tok
  | op  : Op → Tok
deriving Repr

def tokenize : List String → Option (List Tok)
  | [] => some []
  | s :: t =>
    match parseOp s with
    | some o => (tokenize t).map (Tok.op o :: ·)
    | none   => s.toInt?.bind (fun n => (tokenize t).map (Tok.num n :: ·))

partial def evalPass (xs : List Tok) (ops : List Op) : List Tok :=
  match xs with
  | Tok.num a :: Tok.op o :: Tok.num b :: rest =>
    if o ∈ ops then evalPass (Tok.num (apply o a b) :: rest) ops
    else Tok.num a :: Tok.op o :: evalPass (Tok.num b :: rest) ops
  | x :: xs => x :: evalPass xs ops
  | [] => []

def evalTokens (tokens : List Tok) : Option Int :=
  let result := [[.mul, .floordiv], [.add, .sub]].foldl evalPass tokens
  match result with | [Tok.num n] => some n | _ => none

def do_algebra (input : List String) : Option Int :=
  tokenize input >>= evalTokens

    \end{lstlisting}
    \end{minipage}
    % \vspace{-7.3in}
    % (b)\vspace{7.3in}
    \begin{minipage}[t]{0.5\linewidth}
    (b)
    \begin{lstlisting}
def applyOp (x y : Int) : String → Option Int
  | "+"  => some (x + y)
  | "-"  => some (x - y)
  | "*"  => some (x * y)
  | "//" => if y == 0 then none else some (x / y)
  | _     => none

inductive evalArith_pass : List String → Int → Prop
| num {s : String} {n : Nat} (h : s.toNat! = n) :
    evalArith_pass [s] (Int.ofNat n)
| binOp {ts1 ts2 : List String} {op : String} {r1 r2 r : Int}
    (h1 : evalArith_pass ts1 r1)
    (h2 : evalArith_pass ts2 r2)
    (hop : applyOp r1 r2 op = some r) :
    evalArith_pass (ts1 ++ op :: ts2) r

inductive evalArith_mul : List String → Int → Prop
| of_pass {ts r} (h : evalArith_pass ts r) : evalArith_mul ts r
| step {ts1 ts2 r1 r2 r} (h1 : evalArith_mul ts1 r1) (h2 : evalArith_mul ts2 r2)
    (hop : applyOp r1 r2 "*" = some r ∨ applyOp r1 r2 "//" = some r) :
    evalArith_mul (ts1 ++ "*" :: ts2) r

inductive evalArith_add : List String → Int → Prop
| of_mul {ts r} (h : evalArith_mul ts r) : evalArith_add ts r
| step {ts1 ts2 r1 r2 r} (h1 : evalArith_add ts1 r1) (h2 : evalArith_add ts2 r2)
    (hop : applyOp r1 r2 "+" = some r ∨ applyOp r1 r2 "-" = some r) :
    evalArith_add (ts1 ++ "+" :: ts2) r

-- Noncomputable spec to evaluate an expression
def do_algebra (input : List String) (result : Int) : Prop :=
  evalArith_add input result

    \end{lstlisting}
    \end{minipage}
    \end{mdframed}
    %\reducevspacebetweenfigureandcaption
    \vspace{-0.1in}
    \caption{
    Two different specs for evaluating an expression (as a list of strings): \texttt{["2","+","3","*","4","-","5"]}. (a) shows a computable specification that evaluates using \textit{do\_algebra}, and later checked with the result (b) shows a non-computable specification using an inductive definition where \textit{do\_algebra} checks if the result matches the value of the expression without leaks.}
    \label{fig:non-computable-example2}
\end{figure}

\subsection{Baseline Prompts}
\label{sec:baseline-prompts}

Snippets of the few-shot specification generator's system and example prompts are shown in \Cref{fig:specification-generator-system-prompt} and \Cref{fig:specification-generator-example-prompt}. Snippets of the few-shot isomorphism prover's system and example prompts are shown in \Cref{fig:isomorphism-prover-system-prompt} and \Cref{fig:isomorphism-prover-example-prompt}. COPRA's system prompt, used for both isomorphism and correctness, is nearly identical to the original one in the COPRA paper \citep{thakur2024incontext}. Snippets of COPRA's example prompt for isomorphism are shown in \Cref{fig:copra-isomorphism-example-prompt}.

Snippets of the few-shot implementation generator's system and example prompts are shown in \Cref{fig:implementation-generator-system-prompt} and \Cref{fig:implementation-generator-example-prompt}. Snippets of the few-shot correctness prover's system and example prompts are shown in \Cref{fig:correctness-prover-system-prompt} and \Cref{fig:correctness-prover-example-prompt}. Snippets of COPRA's example prompt for correctness are shown in \Cref{fig:copra-correctness-example-prompt}.

\begin{figure}[H]
    \centering
    \begin{mdframed}[roundcorner=10pt]
    \input{text-figures/fig-specification-generator-system-prompt}
    \end{mdframed}
    \vspace{-0.1in}
    \caption{Snippets of the few-shot specification generator's system prompt.}
    \label{fig:specification-generator-system-prompt}
\end{figure}

\begin{figure}[H]
    \centering
    \begin{mdframed}[roundcorner=10pt]
    \input{text-figures/fig-specification-generator-example-prompt}
    \end{mdframed}
    \vspace{-0.1in}
    \caption{Snippets of the few-shot specification generator's example prompt.}
    \label{fig:specification-generator-example-prompt}
\end{figure}

\begin{figure}[H]
    \centering
    \begin{mdframed}[roundcorner=10pt]
    \input{text-figures/fig-isomorphism-prover-system-prompt}
    \end{mdframed}
    \vspace{-0.1in}
    \caption{Snippets of the few-shot isomorphism prover's system prompt.}
    \label{fig:isomorphism-prover-system-prompt}
\end{figure}

\begin{figure}[H]
    \centering
    \begin{mdframed}[roundcorner=10pt]
    \input{text-figures/fig-isomorphism-prover-example-prompt}
    \end{mdframed}
    \vspace{-0.1in}
    \caption{Snippets of the few-shot isomorphism prover's example prompt.}
    \label{fig:isomorphism-prover-example-prompt}
\end{figure}

\begin{figure}[H]
    \centering
    \begin{mdframed}[roundcorner=10pt]
    \input{text-figures/fig-copra-isomorphism-example-prompt}
    \end{mdframed}
    \vspace{-0.1in}
    \caption{Snippets of COPRA's example prompt for isomorphism.}
    \label{fig:copra-isomorphism-example-prompt}
\end{figure}

\begin{figure}[H]
    \centering
    \begin{mdframed}[roundcorner=10pt]
    \input{text-figures/fig-implementation-generator-system-prompt}
    \end{mdframed}
    \vspace{-0.1in}
    \caption{Snippets of the few-shot implementation generator's system prompt.}
    \label{fig:implementation-generator-system-prompt}
\end{figure}

\begin{figure}[H]
    \centering
    \begin{mdframed}[roundcorner=10pt]
    \input{text-figures/fig-implementation-generator-example-prompt}
    \end{mdframed}
    \vspace{-0.1in}
    \caption{Snippets of the few-shot implementation generator's example prompt.}
    \label{fig:implementation-generator-example-prompt}
\end{figure}

\begin{figure}[H]
    \centering
    \begin{mdframed}[roundcorner=10pt]
    \input{text-figures/fig-correctness-prover-system-prompt}
    \end{mdframed}
    \vspace{-0.1in}
    \caption{Snippets of the few-shot correctness prover's system prompt.}
    \label{fig:correctness-prover-system-prompt}
\end{figure}

\begin{figure}[H]
    \centering
    \begin{mdframed}[roundcorner=10pt]
    \input{text-figures/fig-correctness-prover-example-prompt}
    \end{mdframed}
    \vspace{-0.1in}
    \caption{Snippets of the few-shot correctness prover's example prompt.}
    \label{fig:correctness-prover-example-prompt}
\end{figure}

\begin{figure}[H]
    \centering
    \begin{mdframed}[roundcorner=10pt]
    \input{text-figures/fig-copra-correctness-example-prompt}
    \end{mdframed}
    \vspace{-0.1in}
    \caption{Snippets of COPRA's example prompt for correctness.}
    \label{fig:copra-correctness-example-prompt}
\end{figure}

\subsection{Some Proof Found}
\label{sec:app-proofs-found}

\Cref{fig:brazilian-factorial-example} shows an example of a proof found for implementation certification by \texttt{Claude-3.7} using COPRA.

% \begin{wrapfigure}{l}{0.65\textwidth}    
\begin{figure}[t]
\vspace{-0.4in}
\centering
\vspace{-0.05in}
\begin{mdframed}[roundcorner=6pt, innerleftmargin=6pt, innerrightmargin=6pt, innertopmargin=4pt, innerbottommargin=4pt]

\begin{minipage}[t]{1.0\linewidth}
\vspace{-0.25em}
(a)
\begin{lstlisting}[basicstyle=\scriptsize\ttfamily]
def problem_spec (impl : Nat → Nat) (n : Nat) :=
  let spec (result : Nat) :=
    let factorial := Nat.factorial n;
    (0 < n → result / factorial = impl (n - 1)) ∧ 
    (n = 0 → result = 1);
  ∃ result, impl n = result ∧ spec result
\end{lstlisting}
(b)
\begin{lstlisting}[basicstyle=\scriptsize\ttfamily]
def implementation (n : Nat) : Nat :=
  match n with
  | 0 => 1
  | n+1 => Nat.factorial (n+1) * implementation n
\end{lstlisting}
(c)
\begin{lstlisting}[basicstyle=\scriptsize\ttfamily]
#test implementation 4 = 288
\end{lstlisting}
(d)
\begin{lstlisting}[basicstyle=\scriptsize\ttfamily]
theorem correctness (n : Nat) : problem_spec implementation n := by
  unfold problem_spec
  let result := implementation n
  use result
  simp [result]
  apply And.intro
  unfold implementation
  simp [Nat.factorial_succ]
  intro h_n_pos
  cases n
  exfalso
  exact h_n_pos.false
  rename_i n
  simp [Nat.succ_eq_add_one, Nat.add_sub_cancel]
  rw [Nat.factorial_succ]
  have h_fac_pos : 0 < (n + 1) * n.factorial := by
    exact Nat.mul_pos (Nat.succ_pos n) (Nat.factorial_pos n)
  have h_cancel : (n + 1) * n.factorial * implementation n
    / ((n + 1) * n.factorial) = implementation n := by
    rw [Nat.mul_div_cancel_left (implementation n) h_fac_pos]
  simp [h_cancel]
  unfold implementation
  cases n
  simp [Nat.factorial_zero]
  rename_i n
  simp [Nat.add_zero]
  simp [Nat.factorial_succ]
  left
  rw [implementation.eq_def]
  simp [Nat.mul_assoc]
  cases n
  simp [Nat.zero_eq]
  rename_i n
  simp [Nat.factorial_succ]
  rw [Nat.mul_assoc]
  intro h_n_eq_0
  rw [h_n_eq_0, implementation]
\end{lstlisting}
\end{minipage}

\end{mdframed}

\vspace{-0.1in}
\caption{
\textbf{Problem 139 (Brazilian Factorial)}: Given an integer \texttt{n}, compute the product of all factorials from \texttt{n!} down to \texttt{1!}. Part (a) defines the \textbf{ground truth specification}, which expresses recursive structure without leaking the implementation. Part (b) shows the \textbf{implementation} using a recursive product of factorials. Part (c) lists a \textbf{test case} used for validation. Part (d) presents the full \textbf{correctness proof}, showing that the implementation satisfies the spec. This proof, generated by COPRA using Claude-3.7, spans 35 lines and involves reasoning over factorial identities, case analysis, and symbolic manipulation.
}
\label{fig:brazilian-factorial-example}
\end{figure}

\newpage
\newpage
\newpage
\section*{NeurIPS Paper Checklist}

\begin{enumerate}

\item {\bf Claims}
    \item[] Question: Do the main claims made in the abstract and introduction accurately reflect the paper's contributions and scope?
    \item[] Answer: \answerYes{} % Replace by \answerYes{}, \answerNo{}, or \answerNA{}.
    \item[] Justification: Yes, the abstract and introduction discuss our contributions in creating a benchmark for verified code generation.
    \item[] Guidelines:
    \begin{itemize}
        \item The answer NA means that the abstract and introduction do not include the claims made in the paper.
        \item The abstract and/or introduction should clearly state the claims made, including the contributions made in the paper and important assumptions and limitations. A No or NA answer to this question will not be perceived well by the reviewers. 
        \item The claims made should match theoretical and experimental results, and reflect how much the results can be expected to generalize to other settings. 
        \item It is fine to include aspirational goals as motivation as long as it is clear that these goals are not attained by the paper. 
    \end{itemize}

\item {\bf Limitations}
    \item[] Question: Does the paper discuss the limitations of the work performed by the authors?
    \item[] Answer: \answerNA{} % Replace by \answerYes{}, \answerNo{}, or \answerNA{}.
    \item[] Justification: Since this is a benchmark paper, it doesn't have any limitation as such.
    \item[] Guidelines:
    \begin{itemize}
        \item The answer NA means that the paper has no limitation while the answer No means that the paper has limitations, but those are not discussed in the paper. 
        \item The authors are encouraged to create a separate "Limitations" section in their paper.
        \item The paper should point out any strong assumptions and how robust the results are to violations of these assumptions (e.g., independence assumptions, noiseless settings, model well-specification, asymptotic approximations only holding locally). The authors should reflect on how these assumptions might be violated in practice and what the implications would be.
        \item The authors should reflect on the scope of the claims made, e.g., if the approach was only tested on a few datasets or with a few runs. In general, empirical results often depend on implicit assumptions, which should be articulated.
        \item The authors should reflect on the factors that influence the performance of the approach. For example, a facial recognition algorithm may perform poorly when image resolution is low or images are taken in low lighting. Or a speech-to-text system might not be used reliably to provide closed captions for online lectures because it fails to handle technical jargon.
        \item The authors should discuss the computational efficiency of the proposed algorithms and how they scale with dataset size.
        \item If applicable, the authors should discuss possible limitations of their approach to address problems of privacy and fairness.
        \item While the authors might fear that complete honesty about limitations might be used by reviewers as grounds for rejection, a worse outcome might be that reviewers discover limitations that aren't acknowledged in the paper. The authors should use their best judgment and recognize that individual actions in favor of transparency play an important role in developing norms that preserve the integrity of the community. Reviewers will be specifically instructed to not penalize honesty concerning limitations.
    \end{itemize}

\item {\bf Theory assumptions and proofs}
    \item[] Question: For each theoretical result, does the paper provide the full set of assumptions and a complete (and correct) proof?
    \item[] Answer: \answerNA{} % Replace by \answerYes{}, \answerNo{}, or \answerNA{}.
    \item[] Justification: There are no theoretical results discussed in the paper.
    \item[] Guidelines:
    \begin{itemize}
        \item The answer NA means that the paper does not include theoretical results. 
        \item All the theorems, formulas, and proofs in the paper should be numbered and cross-referenced.
        \item All assumptions should be clearly stated or referenced in the statement of any theorems.
        \item The proofs can either appear in the main paper or the supplemental material, but if they appear in the supplemental material, the authors are encouraged to provide a short proof sketch to provide intuition. 
        \item Inversely, any informal proof provided in the core of the paper should be complemented by formal proofs provided in appendix or supplemental material.
        \item Theorems and Lemmas that the proof relies upon should be properly referenced. 
    \end{itemize}

    \item {\bf Experimental result reproducibility}
    \item[] Question: Does the paper fully disclose all the information needed to reproduce the main experimental results of the paper to the extent that it affects the main claims and/or conclusions of the paper (regardless of whether the code and data are provided or not)?
    \item[] Answer: \answerYes{} % Replace by \answerYes{}, \answerNo{}, or \answerNA{}.
    \item[] Justification: All the models used for the experiments are described in the evaluation section. The code used for running these evaluations is shared in the paper.
    \item[] Guidelines:
    \begin{itemize}
        \item The answer NA means that the paper does not include experiments.
        \item If the paper includes experiments, a No answer to this question will not be perceived well by the reviewers: Making the paper reproducible is important, regardless of whether the code and data are provided or not.
        \item If the contribution is a dataset and/or model, the authors should describe the steps taken to make their results reproducible or verifiable. 
        \item Depending on the contribution, reproducibility can be accomplished in various ways. For example, if the contribution is a novel architecture, describing the architecture fully might suffice, or if the contribution is a specific model and empirical evaluation, it may be necessary to either make it possible for others to replicate the model with the same dataset, or provide access to the model. In general. releasing code and data is often one good way to accomplish this, but reproducibility can also be provided via detailed instructions for how to replicate the results, access to a hosted model (e.g., in the case of a large language model), releasing of a model checkpoint, or other means that are appropriate to the research performed.
        \item While NeurIPS does not require releasing code, the conference does require all submissions to provide some reasonable avenue for reproducibility, which may depend on the nature of the contribution. For example
        \begin{enumerate}
            \item If the contribution is primarily a new algorithm, the paper should make it clear how to reproduce that algorithm.
            \item If the contribution is primarily a new model architecture, the paper should describe the architecture clearly and fully.
            \item If the contribution is a new model (e.g., a large language model), then there should either be a way to access this model for reproducing the results or a way to reproduce the model (e.g., with an open-source dataset or instructions for how to construct the dataset).
            \item We recognize that reproducibility may be tricky in some cases, in which case authors are welcome to describe the particular way they provide for reproducibility. In the case of closed-source models, it may be that access to the model is limited in some way (e.g., to registered users), but it should be possible for other researchers to have some path to reproducing or verifying the results.
        \end{enumerate}
    \end{itemize}

\item {\bf Open access to data and code}
    \item[] Question: Does the paper provide open access to the data and code, with sufficient instructions to faithfully reproduce the main experimental results, as described in supplemental material?
    \item[] Answer: \answerYes{} % Replace by \answerYes{}, \answerNo{}, or \answerNA{}.
    \item[] Justification: The code link, data link are shared in the paper. 
    \item[] Guidelines:
    \begin{itemize}
        \item The answer NA means that paper does not include experiments requiring code.
        \item Please see the NeurIPS code and data submission guidelines (\url{https://nips.cc/public/guides/CodeSubmissionPolicy}) for more details.
        \item While we encourage the release of code and data, we understand that this might not be possible, so “No” is an acceptable answer. Papers cannot be rejected simply for not including code, unless this is central to the contribution (e.g., for a new open-source benchmark).
        \item The instructions should contain the exact command and environment needed to run to reproduce the results. See the NeurIPS code and data submission guidelines (\url{https://nips.cc/public/guides/CodeSubmissionPolicy}) for more details.
        \item The authors should provide instructions on data access and preparation, including how to access the raw data, preprocessed data, intermediate data, and generated data, etc.
        \item The authors should provide scripts to reproduce all experimental results for the new proposed method and baselines. If only a subset of experiments are reproducible, they should state which ones are omitted from the script and why.
        \item At submission time, to preserve anonymity, the authors should release anonymized versions (if applicable).
        \item Providing as much information as possible in supplemental material (appended to the paper) is recommended, but including URLs to data and code is permitted.
    \end{itemize}

\item {\bf Experimental setting/details}
    \item[] Question: Does the paper specify all the training and test details (e.g., data splits, hyperparameters, how they were chosen, type of optimizer, etc.) necessary to understand the results?
    \item[] Answer: \answerYes{} % Replace by \answerYes{}, \answerNo{}, or \answerNA{}.
    \item[] Justification: We don't train any model in this work, as it is a benchmark paper. But we mention all the models we have used.
    \item[] Guidelines:
    \begin{itemize}
        \item The answer NA means that the paper does not include experiments.
        \item The experimental setting should be presented in the core of the paper to a level of detail that is necessary to appreciate the results and make sense of them.
        \item The full details can be provided either with the code, in appendix, or as supplemental material.
    \end{itemize}

\item {\bf Experiment statistical significance}
    \item[] Question: Does the paper report error bars suitably and correctly defined or other appropriate information about the statistical significance of the experiments?
    \item[] Answer: \answerNo{} % Replace by \answerYes{}, \answerNo{}, or \answerNA{}.
    \item[] Justification: This is a benchmark paper, we test the capabilities of various LLMs on our problems, these models tend to change from time to time.
    \item[] Guidelines:
    \begin{itemize}
        \item The answer NA means that the paper does not include experiments.
        \item The authors should answer "Yes" if the results are accompanied by error bars, confidence intervals, or statistical significance tests, at least for the experiments that support the main claims of the paper.
        \item The factors of variability that the error bars are capturing should be clearly stated (for example, train/test split, initialization, random drawing of some parameter, or overall run with given experimental conditions).
        \item The method for calculating the error bars should be explained (closed form formula, call to a library function, bootstrap, etc.)
        \item The assumptions made should be given (e.g., Normally distributed errors).
        \item It should be clear whether the error bar is the standard deviation or the standard error of the mean.
        \item It is OK to report 1-sigma error bars, but one should state it. The authors should preferably report a 2-sigma error bar than state that they have a 96\% CI, if the hypothesis of Normality of errors is not verified.
        \item For asymmetric distributions, the authors should be careful not to show in tables or figures symmetric error bars that would yield results that are out of range (e.g. negative error rates).
        \item If error bars are reported in tables or plots, The authors should explain in the text how they were calculated and reference the corresponding figures or tables in the text.
    \end{itemize}

\item {\bf Experiments compute resources}
    \item[] Question: For each experiment, does the paper provide sufficient information on the computer resources (type of compute workers, memory, time of execution) needed to reproduce the experiments?
    \item[] Answer: \answerNo{} % Replace by \answerYes{}, \answerNo{}, or \answerNA{}.
    \item[] Justification: We don't use our own compute as we make calls to models hosted elsewhere, hence we cannot provide an estimate of the amount of compute used.
    \item[] Guidelines:
    \begin{itemize}
        \item The answer NA means that the paper does not include experiments.
        \item The paper should indicate the type of compute workers CPU or GPU, internal cluster, or cloud provider, including relevant memory and storage.
        \item The paper should provide the amount of compute required for each of the individual experimental runs as well as estimate the total compute. 
        \item The paper should disclose whether the full research project required more compute than the experiments reported in the paper (e.g., preliminary or failed experiments that didn't make it into the paper). 
    \end{itemize}
    
\item {\bf Code of ethics}
    \item[] Question: Does the research conducted in the paper conform, in every respect, with the NeurIPS Code of Ethics \url{https://neurips.cc/public/EthicsGuidelines}?
    \item[] Answer: \answerYes{} % Replace by \answerYes{}, \answerNo{}, or \answerNA{}.
    \item[] Justification: I read the code of ethics, and I can assure that my research conforms to it.
    \item[] Guidelines:
    \begin{itemize}
        \item The answer NA means that the authors have not reviewed the NeurIPS Code of Ethics.
        \item If the authors answer No, they should explain the special circumstances that require a deviation from the Code of Ethics.
        \item The authors should make sure to preserve anonymity (e.g., if there is a special consideration due to laws or regulations in their jurisdiction).
    \end{itemize}

\item {\bf Broader impacts}
    \item[] Question: Does the paper discuss both potential positive societal impacts and negative societal impacts of the work performed?
    \item[] Answer: \answerNA{} % Replace by \answerYes{}, \answerNo{}, or \answerNA{}.
    \item[] Justification: Our work has no societal impact because it is mostly about theorem proving.
    \item[] Guidelines:
    \begin{itemize}
        \item The answer NA means that there is no societal impact of the work performed.
        \item If the authors answer NA or No, they should explain why their work has no societal impact or why the paper does not address societal impact.
        \item Examples of negative societal impacts include potential malicious or unintended uses (e.g., disinformation, generating fake profiles, surveillance), fairness considerations (e.g., deployment of technologies that could make decisions that unfairly impact specific groups), privacy considerations, and security considerations.
        \item The conference expects that many papers will be foundational research and not tied to particular applications, let alone deployments. However, if there is a direct path to any negative applications, the authors should point it out. For example, it is legitimate to point out that an improvement in the quality of generative models could be used to generate deepfakes for disinformation. On the other hand, it is not needed to point out that a generic algorithm for optimizing neural networks could enable people to train models that generate Deepfakes faster.
        \item The authors should consider possible harms that could arise when the technology is being used as intended and functioning correctly, harms that could arise when the technology is being used as intended but gives incorrect results, and harms following from (intentional or unintentional) misuse of the technology.
        \item If there are negative societal impacts, the authors could also discuss possible mitigation strategies (e.g., gated release of models, providing defenses in addition to attacks, mechanisms for monitoring misuse, mechanisms to monitor how a system learns from feedback over time, improving the efficiency and accessibility of ML).
    \end{itemize}
    
\item {\bf Safeguards}
    \item[] Question: Does the paper describe safeguards that have been put in place for responsible release of data or models that have a high risk for misuse (e.g., pretrained language models, image generators, or scraped datasets)?
    \item[] Answer: \answerNA{} % Replace by \answerYes{}, \answerNo{}, or \answerNA{}.
    \item[] Justification: Our data is mostly about proving mathematical theorems, hence, should not have any risk involved.
    \item[] Guidelines:
    \begin{itemize}
        \item The answer NA means that the paper poses no such risks.
        \item Released models that have a high risk for misuse or dual-use should be released with necessary safeguards to allow for controlled use of the model, for example by requiring that users adhere to usage guidelines or restrictions to access the model or implementing safety filters. 
        \item Datasets that have been scraped from the Internet could pose safety risks. The authors should describe how they avoided releasing unsafe images.
        \item We recognize that providing effective safeguards is challenging, and many papers do not require this, but we encourage authors to take this into account and make a best faith effort.
    \end{itemize}

\item {\bf Licenses for existing assets}
    \item[] Question: Are the creators or original owners of assets (e.g., code, data, models), used in the paper, properly credited and are the license and terms of use explicitly mentioned and properly respected?
    \item[] Answer: \answerYes{} % Replace by \answerYes{}, \answerNo{}, or \answerNA{}.
    \item[] Justification: All the licenses are mentioned in the GitHub repositories of the links shared.
    \item[] Guidelines:
    \begin{itemize}
        \item The answer NA means that the paper does not use existing assets.
        \item The authors should cite the original paper that produced the code package or dataset.
        \item The authors should state which version of the asset is used and, if possible, include a URL.
        \item The name of the license (e.g., CC-BY 4.0) should be included for each asset.
        \item For scraped data from a particular source (e.g., website), the copyright and terms of service of that source should be provided.
        \item If assets are released, the license, copyright information, and terms of use in the package should be provided. For popular datasets, \url{paperswithcode.com/datasets} has curated licenses for some datasets. Their licensing guide can help determine the license of a dataset.
        \item For existing datasets that are re-packaged, both the original license and the license of the derived asset (if it has changed) should be provided.
        \item If this information is not available online, the authors are encouraged to reach out to the asset's creators.
    \end{itemize}

\item {\bf New assets}
    \item[] Question: Are new assets introduced in the paper well documented and is the documentation provided alongside the assets?
    \item[] Answer: \answerYes{} % Replace by \answerYes{}, \answerNo{}, or \answerNA{}.
    \item[] Justification: All the source code is released, and every Github repository has a README.
    \item[] Guidelines:
    \begin{itemize}
        \item The answer NA means that the paper does not release new assets.
        \item Researchers should communicate the details of the dataset/code/model as part of their submissions via structured templates. This includes details about training, license, limitations, etc. 
        \item The paper should discuss whether and how consent was obtained from people whose asset is used.
        \item At submission time, remember to anonymize your assets (if applicable). You can either create an anonymized URL or include an anonymized zip file.
    \end{itemize}

\item {\bf Crowdsourcing and research with human subjects}
    \item[] Question: For crowdsourcing experiments and research with human subjects, does the paper include the full text of instructions given to participants and screenshots, if applicable, as well as details about compensation (if any)? 
    \item[] Answer: \answerNA{} % Replace by \answerYes{}, \answerNo{}, or \answerNA{}.
    \item[] Justification: Our paper does not involve crowdsourcing nor research with human subjects.
    \item[] Guidelines:
    \begin{itemize}
        \item The answer NA means that the paper does not involve crowdsourcing nor research with human subjects.
        \item Including this information in the supplemental material is fine, but if the main contribution of the paper involves human subjects, then as much detail as possible should be included in the main paper. 
        \item According to the NeurIPS Code of Ethics, workers involved in data collection, curation, or other labor should be paid at least the minimum wage in the country of the data collector. 
    \end{itemize}

\item {\bf Institutional review board (IRB) approvals or equivalent for research with human subjects}
    \item[] Question: Does the paper describe potential risks incurred by study participants, whether such risks were disclosed to the subjects, and whether Institutional Review Board (IRB) approvals (or an equivalent approval/review based on the requirements of your country or institution) were obtained?
    \item[] Answer: \answerNA{} % Replace by \answerYes{}, \answerNo{}, or \answerNA{}.
    \item[] Justification: Our paper does not involve crowdsourcing nor research with human subjects.
    \item[] Guidelines:
    \begin{itemize}
        \item The answer NA means that the paper does not involve crowdsourcing nor research with human subjects.
        \item Depending on the country in which research is conducted, IRB approval (or equivalent) may be required for any human subjects research. If you obtained IRB approval, you should clearly state this in the paper. 
        \item We recognize that the procedures for this may vary significantly between institutions and locations, and we expect authors to adhere to the NeurIPS Code of Ethics and the guidelines for their institution. 
        \item For initial submissions, do not include any information that would break anonymity (if applicable), such as the institution conducting the review.
    \end{itemize}

\item {\bf Declaration of LLM usage}
    \item[] Question: Does the paper describe the usage of LLMs if it is an important, original, or non-standard component of the core methods in this research? Note that if the LLM is used only for writing, editing, or formatting purposes and does not impact the core methodology, scientific rigorousness, or originality of the research, declaration is not required.
    %this research? 
    \item[] Answer: \answerNA{} % Replace by \answerYes{}, \answerNo{}, or \answerNA{}.
    \item[] Justification: The core method development was not carried out with the involvement of LLMs.
    \item[] Guidelines:
    \begin{itemize}
        \item The answer NA means that the core method development in this research does not involve LLMs as any important, original, or non-standard components.
        \item Please refer to our LLM policy (\url{https://neurips.cc/Conferences/2025/LLM}) for what should or should not be described.
    \end{itemize}

\end{enumerate}

\end{document}